\documentclass{article}

\usepackage[sort&compress]{natbib}

\usepackage[final]{neurips_2021}

\usepackage[utf8]{inputenc}
\usepackage[T1]{fontenc}
\usepackage{url, booktabs, amsfonts, nicefrac, microtype, xcolor}
\usepackage{bm, amsmath, amssymb, siunitx, tcolorbox, pifont, xparse}
\usepackage{tabulary, multirow, overpic, xspace, rotating, subcaption}
\usepackage[pagebackref=true, breaklinks=true, letterpaper=true, colorlinks,
            citecolor=citecolor, bookmarks=false]{hyperref}
\setcitestyle{square,comma,numbers}
\definecolor{citecolor}{RGB}{34,139,34}

\xspaceaddexceptions{]\}}
\renewcommand{\paragraph}[1]{\vspace{1.25mm}\noindent\textbf{#1}}
\newcommand{\tablestyle}[2]{\setlength{\tabcolsep}{#1}\renewcommand{\arraystretch}{#2}\centering\footnotesize}
\newlength\savewidth\newcommand\shline{\noalign{\global\savewidth\arrayrulewidth
  \global\arrayrulewidth 1pt}\hline\noalign{\global\arrayrulewidth\savewidth}}
\makeatletter
\DeclareRobustCommand\onedot{\futurelet\@let@token\@onedot}
\def\@onedot{\ifx\@let@token.\else.\null\fi\xspace}
\def\eg{\emph{e.g}\onedot} \def\Eg{\emph{E.g}\onedot}
\def\ie{\emph{i.e}\onedot} 
\def\cf{\emph{cf}\onedot} 
\def\etc{\emph{etc}\onedot} \def\vs{\emph{vs}\onedot}
\def\wrt{w.r.t\onedot} 
\def\etal{\emph{et al}\onedot}
\makeatother

\newcommand{\aff}[1]{${\textstyle^{#1}}$}

\newcommand{\app}{\raise.17ex\hbox{$\scriptstyle\sim$}}
\newcommand{\expnumber}[2]{{#1}\mathrm{e}{#2}}
\newcommand{\x}{{\times}}
\newcommand{\lr}{\emph{lr}\xspace}
\newcommand{\wtd}{\emph{wd}\xspace}
\NewDocumentCommand\vit{mg}{{ViT$_{#1}$\IfNoValueF{#2}{-#2}}}
\newcommand{\cmark}{\ding{51}\xspace}%
\newcommand{\ph}{\phantom{0}}
\newcommand{\phh}{\phantom{00}}
\definecolor{Gray}{gray}{0.6}
\newcommand{\gc}[1]{\textcolor{Gray}{#1}}
\newcommand{\bc}[1]{\textcolor{blue}{#1}}

\title{Early Convolutions Help Transformers See Better}

\author{%
Tete Xiao\aff{1,2}\hspace{1.4mm}%
Mannat Singh\aff{1}\hspace{1.4mm}%
Eric Mintun\aff{1}\hspace{1.4mm}%
Trevor Darrell\aff{2}\hspace{1.4mm}%
Piotr Doll\'ar\aff{1\ast}\hspace{1.4mm}%
Ross Girshick\aff{1\ast}\\\\%
\aff{1}Facebook AI Research (FAIR)\hspace{10mm}\aff{2}UC Berkeley%
}

\begin{document}

\maketitle

\begin{abstract}
Vision transformer (ViT) models exhibit substandard optimizability. In particular, they are sensitive to the choice of optimizer (AdamW \vs SGD), optimizer hyperparameters, and training schedule length. In comparison, modern convolutional neural networks are easier to optimize. Why is this the case? In this work, we conjecture that the issue lies with the \emph{patchify stem} of ViT models, which is implemented by a stride-$p$ $p \x p$ convolution ($p=16$ by default) applied to the input image. This large-kernel plus large-stride convolution runs counter to typical design choices of convolutional layers in neural networks. To test whether this atypical design choice causes an issue, we analyze the optimization behavior of ViT models with their original patchify stem versus a simple counterpart where we replace the ViT stem by a small number of stacked stride-two $3\x3$ convolutions. While the vast majority of computation in the two ViT designs is identical, we find that this small change in early visual processing results in markedly different training behavior in terms of the sensitivity to optimization settings as well as the final model accuracy. Using a \emph{convolutional stem} in ViT dramatically increases optimization stability and also improves peak performance (by \app1-2\% top-1 accuracy on ImageNet-1k), while maintaining flops and runtime. The improvement can be observed across the wide spectrum of model complexities (from 1G to 36G flops) and dataset scales (from ImageNet-1k to ImageNet-21k). These findings lead us to recommend using a standard, lightweight convolutional stem for ViT models in this regime as a more robust architectural choice compared to the original ViT model design.
\end{abstract}

\section{Introduction}\label{sec:intro}

\begin{figure}[t]\centering
\includegraphics[width=1.0\linewidth]{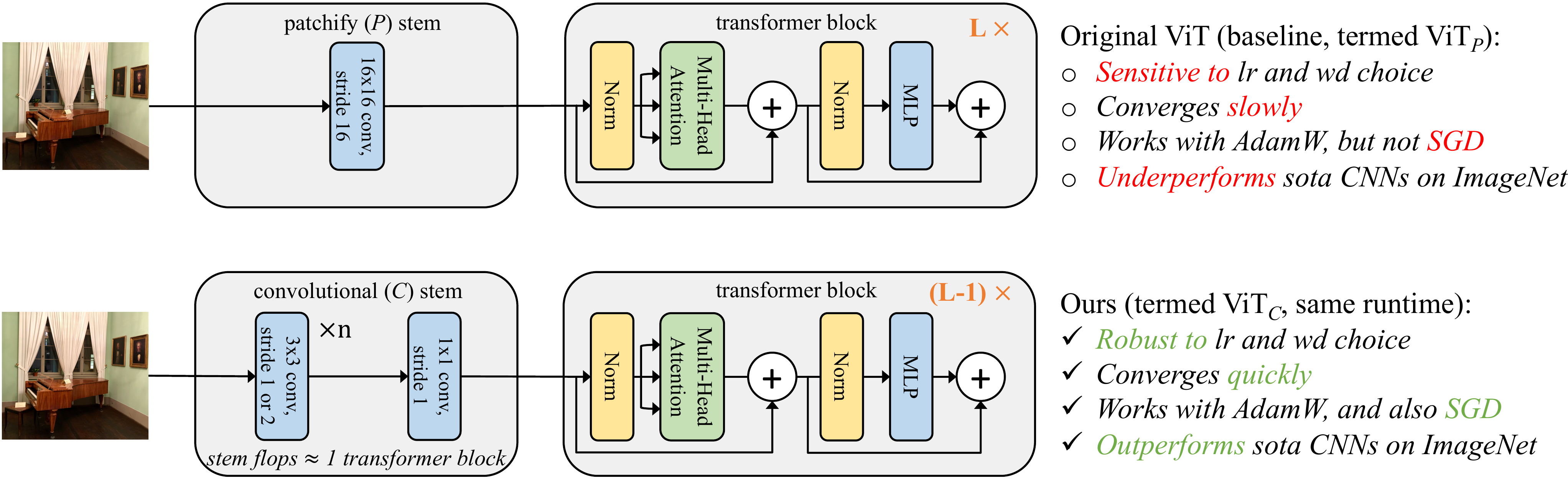}
\caption{\textbf{Early convolutions help transformers see better}: We hypothesize that the substandard optimizability of ViT models compared to CNNs primarily arises from the \emph{early} visual processing performed by its \emph{patchify stem}, which is implemented by a non-overlapping stride-$p$ $p\x p$ convolution, with $p=16$ by default. We \emph{minimally} replace the patchify stem in ViT with a standard \emph{convolutional stem} of only \app5 convolutions that has approximately the same complexity as a \emph{single} transformer block. We reduce the number of transformer blocks by one (\ie, $L-1$ \vs $L$) to maintain parity in flops, parameters, and runtime. We refer to the resulting model as \vit{C} and the original \vit{} as \vit{P}. The vast majority of computation performed by these two models is identical, yet surprisingly we observe that \vit{C} (i) converges faster, (ii) enables, for the first time, the use of either AdamW or SGD without a significant accuracy drop, (iii) shows greater stability to learning rate and weight decay choice, and (iv) yields improvements in ImageNet top-1 error allowing \vit{C} to outperform state-of-the-art CNNs, whereas \vit{P} does not.}
\label{fig:teaser}
\vspace{-3mm}
\end{figure}

Vision transformer (ViT) models~\cite{Dosovitskiy2020image} offer an alternative design paradigm to convolutional neural networks (CNNs)~\cite{LeCun1989}. ViTs replace the inductive bias towards local processing inherent in convolutions with global processing performed by multi-headed self-attention~\cite{Vaswani2017attention}. The hope is that this design has the potential to improve performance on vision tasks, akin to the trends observed in natural language processing~\cite{devlin2019bert}. While investigating this conjecture, researchers face another unexpected difference between ViTs and CNNs: ViT models exhibit substandard \emph{optimizability}. ViTs are sensitive to the choice of optimizer~\cite{Touvron2020training} (AdamW~\cite{Loshchilov2017decoupled} \vs SGD), to the selection of dataset specific learning hyperparameters~\cite{Dosovitskiy2020image,Touvron2020training}, to training schedule length, to network depth~\cite{touvron2021going}, \etc. These issues render former training recipes and intuitions ineffective and impede research.

Convolutional neural networks, in contrast, are exceptionally easy and robust to optimize. Simple training recipes based on SGD, basic data augmentation, and standard hyperparameter values have been widely used for years~\cite{He2016}. Why does this difference exist between ViT and CNN models? In this paper we hypothesize that the issues lies primarily in the \emph{early} visual processing performed by ViT. ViT ``patchifies'' the input image into $p\x p$ non-overlapping patches to form the transformer encoder's input set. This \emph{patchify stem} is implemented as a stride-$p$ $p\x p$ convolution, with $p=16$ as a default value. This large-kernel plus large-stride convolution runs counter to the typical design choices used in CNNs, where best-practices have converged to a small stack of stride-two $3\x 3$ kernels as the network's stem (\eg, ~\cite{Simonyan2015,Tan2019,Radosavovic2019}).

To test this hypothesis, we \emph{minimally} change the early visual processing of ViT by replacing its patchify stem with a standard \emph{convolutional stem} consisting of only $\app5$ convolutions, see Figure~\ref{fig:teaser}. To compensate for the small addition in flops, we remove one transformer block to maintain parity in flops and runtime. We observe that even though the vast majority of the computation in the two ViT designs is identical, this small change in early visual processing results in markedly different training behavior in terms of the sensitivity to optimization settings as well as the final model accuracy.

In extensive experiments we show that replacing the ViT patchify stem with a more standard convolutional stem (i) allows ViT to converge faster (\S\ref{sec:experiments:length}), (ii) enables, for the first time, the use of either AdamW or SGD without a significant drop in accuracy (\S\ref{sec:experiments:optimizer}), (iii) brings ViT's stability \wrt learning rate and weight decay closer to that of modern CNNs (\S\ref{sec:experiments:lrwd}), and (iv) yields improvements in ImageNet~\cite{Deng2009} top-1 error of \app 1-2 percentage points (\S\ref{sec:reality}). We consistently observe these improvements across a wide spectrum of model complexities (from 1G flops to 36G flops) and dataset scales (ImageNet-1k to ImageNet-21k).

These results show that injecting some convolutional inductive bias into ViTs can be beneficial under commonly studied settings. We did \emph{not} observe evidence that the hard locality constraint in early layers hampers the representational capacity of the network, as might be feared~\cite{dascoli2021convit}. In fact we observed the opposite, as ImageNet results improve even with larger-scale models and larger-scale data when using a convolution stem. Moreover, under carefully controlled comparisons, we find that ViTs are only able to surpass state-of-the-art CNNs when equipped with a convolutional stem (\S\ref{sec:reality}).

We conjecture that restricting convolutions in ViT to \emph{early} visual processing may be a crucial design choice that strikes a balance between (hard) inductive biases and the representation learning ability of transformer blocks. Evidence comes by comparison to the ``hybrid ViT'' presented in~\cite{Dosovitskiy2020image}, which uses 40 convolutional layers (most of a ResNet-50) and shows no improvement over the default ViT. This perspective resonates with the findings of~\cite{dascoli2021convit}, who observe that early transformer blocks prefer to learn more local attention patterns than later blocks. Finally we note that exploring the design of hybrid CNN/ViT models is \emph{not} a goal of this work; rather we demonstrate that simply using a minimal convolutional stem with ViT is sufficient to dramatically change its optimization behavior.

In summary, the findings presented in this paper lead us to recommend using a standard, lightweight convolutional stem for ViT models in the analyzed dataset scale and model complexity spectrum as a more robust and higher performing architectural choice compared to the original ViT model design.

\section{Related Work}\label{sec:relatedwork}

\paragraph{Convolutional neural networks (CNNs).} The breakthrough performance of the AlexNet~\cite{Krizhevsky2012} CNN~\cite{Fukushima1980neocognitron,LeCun1989} on ImageNet classification~\cite{Deng2009} transformed the field of recognition, leading to the development of higher performing architectures, \eg, \cite{Simonyan2015,Szegedy2015,He2016,Xie2017}, and scalable training methods~\cite{Ioffe2015,Goyal2017}. These architectures are now core components in object detection (\eg,~\cite{Ren2015}), instance segmentation (\eg,~\cite{He2017}), and semantic segmentation (\eg,~\cite{Long2015fully}). CNNs are typically trained with stochastic gradient descent (SGD) and are widely considered to be easy to optimize.

\paragraph{Self-attention in vision models.} Transformers~\cite{Vaswani2017attention} are revolutionizing natural language processing by enabling scalable training. Transformers use multi-headed self-attention, which performs global information processing and is strictly more general than convolution~\cite{Cordonnier2020relationship}. Wang \etal~\cite{Wang2018non} show that (single-headed) self-attention is a form of non-local means~\cite{Buades2005non} and that integrating it into a ResNet~\cite{He2016} improves several tasks. Ramachandran \etal~\cite{Ramachandran2019stand} explore this direction further with stand-alone self-attention networks for vision. They report difficulties in designing an attention-based network stem and present a bespoke solution that avoids convolutions. In contrast, we demonstrate the benefits of a convolutional stem. Zhao \etal~\cite{Zhao2020exploring} explore a broader set of self-attention operations with hard-coded locality constraints, more similar to standard CNNs.

\paragraph{Vision transformer (ViT).} Dosovitskiy \etal~\cite{Dosovitskiy2020image} apply a transformer encoder to image classification with minimal vision-specific modifications. As the counterpart of input token embeddings, they partition the input image into, \eg, $16\x16$ pixel, non-overlapping patches and linearly project them to the encoder's input dimension. They report lackluster results when training on ImageNet-1k, but demonstrate state-of-the-art transfer learning when using large-scale pretraining data. ViTs are sensitive to many details of the training recipe, \eg, they benefit greatly from AdamW~\cite{Loshchilov2017decoupled} compared to SGD and require careful learning rate and weight decay selection. ViTs are generally considered to be difficult to optimize compared to CNNs (\eg, see~\cite{Dosovitskiy2020image,Touvron2020training,touvron2021going}). Further evidence of challenges comes from Chen \etal~\cite{chen2021empirical} who report ViT optimization instability in self-supervised learning (unlike with CNNs), and find that freezing the patchify stem at its random initialization improves stability.

\paragraph{ViT improvements.} ViTs are gaining rapid interest in part because they may offer a novel direction away from CNNs. Touvron \etal~\cite{Touvron2020training} show that with more regularization and stronger data augmentation ViT models achieve competitive accuracy on ImageNet-1k alone (\cf~\cite{Dosovitskiy2020image}). Subsequently, works concurrent with our own explore numerous other ViT improvements. Dominant themes include multi-scale networks~\cite{graham2021levit,fan2021multiscale,yuan2021tokens,wang2021pyramid,liu2021swin}, increasing depth~\cite{touvron2021going}, and locality priors~\cite{dascoli2021convit,chen2021visformer,wu2021cvt,yuan2021incorporating,graham2021levit}. In~\cite{dascoli2021convit}, d'Ascoli \etal modify multi-head self-attention with a convolutional bias at initialization and show that this prior improves sample efficiency and ImageNet accuracy. Resonating with our work, \cite{chen2021visformer,wu2021cvt,yuan2021incorporating,graham2021levit} present models with convolutional stems, but do not analyze optimizability (our focus).

\paragraph{Discussion.} Unlike the concurrent work on locality priors in ViT, our focus is studying \emph{optimizability} under \emph{minimal} ViT modifications in order to derive crisp conclusions. Our perspective brings several novel observations: by adding only \app 5 convolutions to the stem, ViT can be optimized well with either AdamW or SGD (\cf~all prior works use AdamW to avoid large drops in accuracy~\cite{Touvron2020training}), it becomes less sensitive to the specific choice of learning rate and weight decay, and training converges faster. We also observe a consistent improvement in ImageNet top-1 accuracy across a wide spectrum of model complexities (1G flops to 36G flops) and dataset scales (ImageNet-1k to ImageNet-21k). These results suggest that a (hard) convolutional bias early in the network does not compromise representational capacity, as conjectured in~\cite{dascoli2021convit}, and is beneficial within the scope of this study.

\section{Vision Transformer Architectures}\label{sec:architectures}

Next, we review vision transformers~\cite{Dosovitskiy2020image} and describe the convolutional stems used in our work.

\paragraph{The vision transformer (ViT).} ViT first partitions an input image into \emph{non-overlapping} $p \x p$ patches and linearly projects each patch to a $d$-dimensional feature vector using a learned weight matrix. A patch size of $p=16$ and an image size of $224\x224$ are typical. The resulting patch embeddings (plus positional embeddings and a learned classification token embedding) are processed by a standard transformer encoder~\cite{Vaswani2017attention,wang2019learning} followed by a classification head. Using common network nomenclature, we refer to the portion of ViT before the transformer blocks as the network's \emph{stem}. ViT's stem is a specific case of convolution (stride-$p$, $p\x p$ kernel), but we will refer to it as the \emph{patchify stem} and reserve the terminology of \emph{convolutional stem} for stems with a more conventional CNN design with multiple layers of \emph{overlapping} convolutions (\ie, with stride smaller than the kernel size).

\paragraph{\vit{P} models.} Prior work proposes ViT models of various sizes, such as ViT-Tiny, ViT-Small, ViT-Base, \etc~\cite{Dosovitskiy2020image,Touvron2020training}. To facilitate comparisons with CNNs, which are typically standardized to 1 gigaflop (GF), 2GF, 4GF, 8GF, \etc, we modify the original ViT models to obtain models at about these complexities. Details are given in Table~\ref{tab:architectures:encoders_and_complexity} (left). For easier comparison with CNNs of similar flops, and to avoid subjective size names, we refer the models by their flops, \eg, \vit{P}{4GF} in place of ViT-Small. We use the $P$ subscript to indicate that these models use the original \emph{patchify} stem.

\begin{table*}[t]\centering\vspace{-1mm}
\tablestyle{2pt}{1.1}\small
\resizebox{0.510\textwidth}{!}{\centering
\begin{tabular}{@{}l|c|cccc|cccc@{}}
\multirow{2}{*}{model} & ref & hidden & MLP & num & num & flops & params & acts & time \\
& model & size & mult & heads & blocks & (B) & (M) & (M) & (min) \\\shline
\vit{P}{1GF} & \app ViT-T & 192 & \bc{3} & 3 & 12 & 1.1 & 4.8 & 5.5 & 2.6 \\
\vit{P}{4GF} & \app ViT-S & 384 & \bc{3} & 6 & 12 & 3.9 & 18.5 & 11.1 & 3.8 \\
\vit{P}{18GF} & =ViT-B & 768 & 4 & 12 & 12 & 17.5 & 86.7 & 24.0 & 11.5 \\
\vit{P}{36GF} & $\frac{3}{5}$ViT-L & 1024 & 4 & 16 & \bc{14} & 35.9 & 178.4 & 37.3 & 18.8 \\
\end{tabular}}
\hfill
\resizebox{0.455\textwidth}{!}{\centering
\begin{tabular}{@{}l|cccc|cccc@{}}
\multirow{2}{*}{model} & hidden & MLP & num & num & flops & params & acts & time \\
& size & mult & heads & blocks & (B) & (M) & (M) & (min) \\\shline
\vit{C}{1GF} & 192 & 3 & 3 & \bc{11} & 1.1 & 4.6 & 5.7 & 2.7 \\
\vit{C}{4GF} & 384 & 3 & 6 & \bc{11} & 4.0 & 17.8 & 11.3 & 3.9 \\
\vit{C}{18GF} & 768 & 4 & 12 & \bc{11} & 17.7 & 81.6 & 24.1 & 11.4 \\
\vit{C}{36GF} & 1024 & 4 & 16 & \bc{13} & 35.0 & 167.8 & 36.7 & 18.6 \\
\end{tabular}}
\vspace{1.5mm}
\caption{\textbf{Model definitions}: \textit{Left}: Our \vit{P} models at various complexities, which use the original \emph{patchify} stem and closely resemble the original ViT models~\cite{Dosovitskiy2020image}. To facilitate comparisons with CNNs, we modify the original ViT-Tiny, -Small, -Base, -Large models to obtain models at 1GF, 4GF, 18GF, and 36GF, respectively. The modifications are indicated in blue and include reducing the MLP multiplier from 4$\x$ to 3$\x$ for the 1GF and 4GF models, and reducing the number of transformer blocks from 24 to 14 for the 36GF model. \textit{Right}: Our \vit{C} models at various complexities that use the \emph{convolutional} stem. The only additional modification relative to the corresponding \vit{P} models is the removal of 1 transformer block to compensate for the increased flops of the convolutional stem. We show complexity measures for all models (flops, parameters, activations, and epoch training time on ImageNet-1k); the corresponding \vit{P} and \vit{C} models match closely on all metrics.}
\label{tab:architectures:encoders_and_complexity}
\vspace{-4mm}
\end{table*}

\paragraph{Convolutional stem design.} We adopt a typical minimalist convolutional stem design by stacking $3\x3$ convolutions~\cite{Simonyan2015}, followed by a single $1\x1$ convolution at the end to match the $d$-dimensional input of the transformer encoder. These stems quickly downsample a $224\x224$ input image using overlapping strided convolutions to $14\x14$, matching the number of inputs created by the standard patchify stem. We follow a simple design pattern: all $3\x3$ convolutions either have stride 2 and double the number of output channels or stride 1 and keep the number of output channels constant. We enforce that the stem accounts for approximately the computation of one transformer block of the corresponding model so that we can easily control for flops by removing one transformer block when using the convolutional stem instead of the patchify stem. Our stem design was chosen to be purposefully simple and we emphasize that it was not designed to maximize model accuracy.

\paragraph{\vit{C} models.} To form a ViT model with a convolutional stem, we simply replace the patchify stem with its counterpart convolutional stem and \emph{remove one transformer block} to compensate for the convolutional stem's extra flops (see Figure~\ref{fig:teaser}). We refer to the modified ViT with a convolutional stem as \vit{C}. Configurations for \vit{C} at various complexities are given in Table~\ref{tab:architectures:encoders_and_complexity} (right); corresponding \vit{P} and \vit{C} models match closely on all complexity metrics including flops and runtime.

\paragraph{Convolutional stem details.} Our convolutional stem designs use four, four, and six $3\x3$ convolutions for the 1GF, 4GF, and 18GF models, respectively. The output channels are [24, 48, 96, 192], [48, 96, 192, 384], and [64, 128, 128, 256, 256, 512], respectively. All $3\x3$ convolutions are followed by batch norm (BN)~\cite{Ioffe2015} and then ReLU~\cite{Nair2010}, while the final $1\x1$ convolution is not, to be consistent with the original patchify stem. Eventually, matching stem flops to transformer block flops results in an unreasonably large stem, thus \vit{C}{36GF} uses the same stem as \vit{C}{18GF}.

\paragraph{Convolutions in ViT.} Dosovitskiy~\etal~\cite{Dosovitskiy2020image} also introduced a ``hybrid ViT'' architecture that blends a modified ResNet~\cite{He2016} (BiT-ResNet~\cite{kolesnikov2019big}) with a transformer encoder. In their hybrid model, the patchify stem is replaced by a partial BiT-ResNet-50 that terminates at the output of the conv4 stage or the output of an extended conv3 stage. These image embeddings replace the standard patchify stem embeddings. This partial BiT-ResNet-50 stem is \emph{deep}, with 40 convolutional layers. In this work, we explore \emph{lightweight} convolutional stems that consist of only 5 to 7 convolutions in total, instead of the 40 used by the hybrid ViT. Moreover, we emphasize that the goal of our work is \emph{not} to explore the hybrid ViT design space, but rather to study the optimizability effects of simply replacing the patchify stem with a \emph{minimal} convolutional stem that follows standard CNN design practices.

\section{Measuring Optimizability}\label{sec:stability}

It has been noted in the literature that ViT models are challenging to optimize, \eg, they may achieve only modest performance when trained on a mid-size dataset (ImageNet-1k)~\cite{Dosovitskiy2020image}, are sensitive to data augmentation~\cite{Touvron2020training} and optimizer choice~\cite{Touvron2020training}, and may perform poorly when made deeper~\cite{touvron2021going}. We empirically observed the general presence of such difficulties through the course of our experiments and informally refer to such optimization characteristics collectively as \emph{optimizability}.

Models with poor optimizability can yield very different results when hyperparameters are varied, which can lead to seemingly bizarre observations, \eg, removing \emph{erasing} data augmentation~\cite{zhong2020random} causes a catastrophic drop in ImageNet accuracy in~\cite{Touvron2020training}. Quantitative metrics to measure optimizability are needed to allow for more robust comparisons. In this section, we establish the foundations of such comparisons; we extensively test various models using these optimizability measures in \S\ref{sec:experiments}.

\paragraph{Training length stability.} Prior works train ViT models for lengthy schedules, \eg, 300 to 400 epochs on ImageNet is typical (at the extreme, \cite{graham2021levit} trains models for 1000 epochs), since results at a formerly common 100-epoch schedule are substantially worse (2-4\% lower top-1 accuracy, see \S\ref{sec:experiments:length}). In the context of ImageNet, we define top-1 accuracy at 400 epochs as an approximate asymptotic result, \ie, training for longer will not meaningfully improve top-1 accuracy, and we compare it to the accuracy of models trained for only 50, 100, or 200 epochs. We define \emph{training length stability} as the gap to asymptotic accuracy. Intuitively, it's a measure of convergence speed. Models that converge faster offer obvious practical benefits, especially when training many model variants.

\paragraph{Optimizer stability.} Prior works use AdamW~\cite{Loshchilov2017decoupled} to optimize ViT models from random initialization. Results of SGD are not typically presented and we are only aware of Touvron \etal~\cite{Touvron2020training}'s report of a dramatic \app 7\% drop in ImageNet top-1 accuracy. In contrast, widely used CNNs, such as ResNets, can be optimized equally well with either SGD or AdamW (see~\S\ref{sec:experiments:optimizer}) and SGD (always with momentum) is typically used in practice. SGD has the practical benefit of having fewer hyperparameters (\eg, tuning AdamW's $\beta_2$ can be important~\cite{chen2020generative}) and requiring 50\% less optimizer state memory, which can ease scaling. We define \emph{optimizer stability} as the accuracy gap between AdamW and SGD. Like training length stability, we use optimizer stability as a proxy for the ease of optimization of a model.

\paragraph{Hyperparameter (\lr, \wtd) stability.} Learning rate (\lr) and weight decay (\wtd) are among the most important hyperparameters governing optimization with SGD and AdamW. New models and datasets often require a search for their optimal values as the choice can dramatically affect results. It is desirable to have a model and optimizer that yield good results for a wide range of learning rate and weight decay values. We will explore this \emph{hyperparameter stability} by comparing the error distribution functions (EDFs)~\cite{Radosavovic2019} of models trained with various choices of \lr and \wtd. In this setting, to create an EDF for a model we randomly sample values of \lr and \wtd and train the model accordingly. Distributional estimates, like those provided by EDFs, give a more complete view of the characteristics of models that point estimates cannot reveal~\cite{Radosavovic2019,Radosavovic2020}. We will review EDFs in \S\ref{sec:experiments:lrwd}.

\paragraph{Peak performance.} The maximum possible performance of each model is the most commonly used metric in previous literature and it is often provided without carefully controlling training details such as data augmentations, regularization methods, number of epochs, and \lr, \wtd tuning. To make more robust comparisons, we define \emph{peak performance} as the result of a model at 400 epochs using its best-performing optimizer and \emph{parsimoniously} tuned \lr and \wtd values (details in \S\ref{sec:reality}), \emph{while fixing justifiably good values for all other variables that have a known impact on training}. Peak performance results for ViTs and CNNs under these carefully controlled training settings are presented in \S\ref{sec:reality}.

\section{Stability Experiments}\label{sec:experiments}

In this section we test the \emph{stability} of ViT models with the original patchify ($P$) stem \vs the convolutional ($C$) stem defined in~\S\ref{sec:architectures}. For reference, we also train RegNetY~\cite{Radosavovic2020,dollar2021fast}, a state-of-the-art CNN that is easy to optimize and serves as a reference point for good stability.

We conduct experiments using ImageNet-1k~\cite{Deng2009}'s standard training and validation sets, and report top-1 error. Following \cite{dollar2021fast}, for all results, we carefully control training settings and we use a minimal set of data augmentations that still yields strong results, for details see \S\ref{sec:experiments:details}. In this section, unless noted, for each model we use the optimal \lr and \wtd found under a 50 epoch schedule (see Appendix).

\begin{figure}[t]\centering
\includegraphics[width=\linewidth]{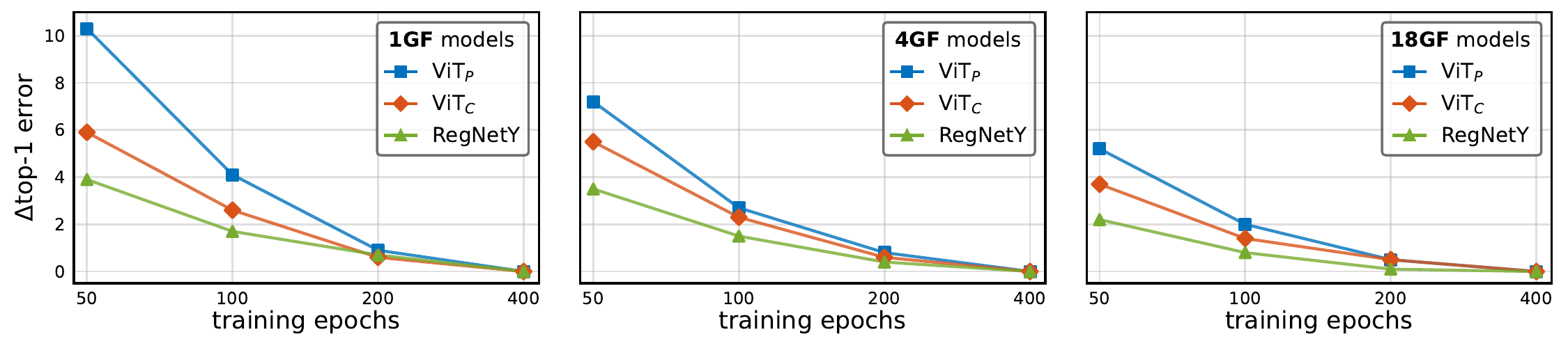}
\caption{\textbf{Training length stability}: We train 9 models for 50 to 400 epochs on ImageNet-1k and plot the $\Delta$top-1 error to the 400 epoch result for each. \vit{C} demonstrates faster convergence than \vit{P} across the model complexity spectrum, and helps close the gap to CNNs (represented by RegNetY).}
\label{fig:experiments:length}
\end{figure}

\subsection{Training Length Stability}\label{sec:experiments:length}

We first explore how rapidly networks converge to their asymptotic error on ImageNet-1k, \ie, the highest possible accuracy achievable by training for many epochs. We approximate asymptotic error as a model's error using a 400 epoch schedule based on observing diminishing returns from 200 to 400. We consider a grid of 24 experiments for \vit{}: \{$P$, $C$\} stems $\x$ \{1, 4, 18\} GF model sizes $\x$ \{50, 100, 200, 400\} epochs. For reference we also train RegNetY at \{1, 4, 16\} GF. We use the best optimizer choice for each model (AdamW for ViT models and SGD for RegNetY models).

\paragraph{Results.} Figure~\ref{fig:experiments:length} shows the absolute error \emph{deltas} ($\Delta$top-1) between 50, 100, and 200 epoch schedules and asymptotic performance (at 400 epochs). \vit{C} demonstrates faster convergence than \vit{P} across the model complexity spectrum, and closes much of the gap to the rate of CNN convergence. The improvement is most significant in the shortest training schedule (50 epoch), \eg, \vit{P}-1GF has a 10\% error delta, while \vit{C}-1GF reduces this to about 6\%. This opens the door to applications that execute a large number of short-scheduled experiments, such as neural architecture search.

\begin{figure}[t]\centering
\includegraphics[width=1.0\linewidth]{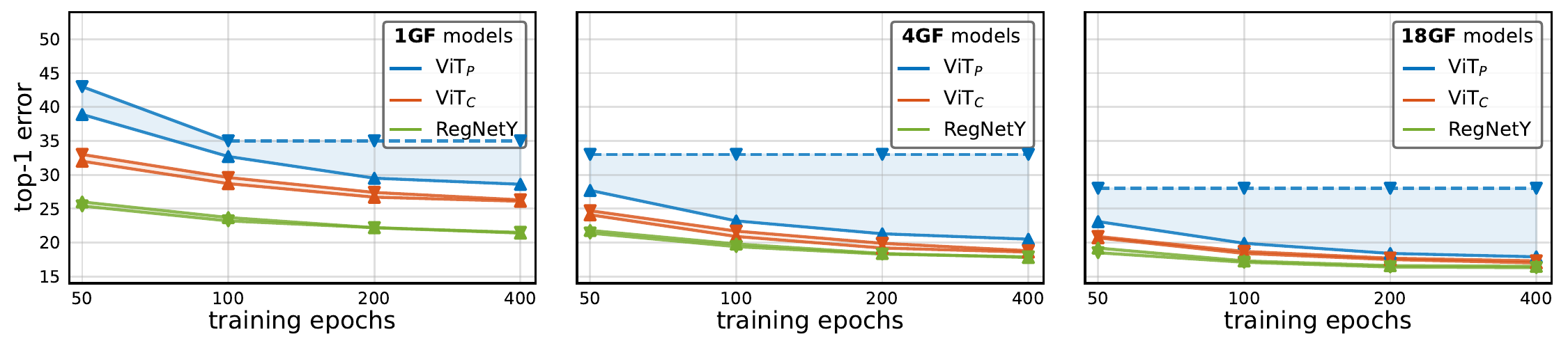}
\caption{\textbf{Optimizer stability}: We train each model for 50 to 400 epochs with AdamW (upward triangle {\small $\blacktriangle$}) and SGD (downward triangle {\small $\blacktriangledown$}). For the baseline \vit{P}, SGD yields significantly worse results than AdamW. In contrast, \vit{C} and RegNetY models exhibit a much smaller gap between SGD and AdamW across all settings. Note that for long schedules, \vit{P} often fails to converge with SGD (\ie, loss goes to NaN), in such cases we copy the best results from a shorter schedule of the same model (and show the results via a dashed line).}
\label{fig:experiments:optimizer}
\end{figure}

\subsection{Optimizer Stability}\label{sec:experiments:optimizer}

We next explore how well AdamW and SGD optimize ViT models with the two stem types. We consider the following grid of 48 \vit{} experiments: \{$P$, $C$\} stems $\x$ \{1, 4, 18\} GF sizes $\x$ \{50, 100, 200, 400\} epochs $\x$ \{AdamW, SGD\} optimizers. As a reference, we also train 24 RegNetY baselines, one for each complexity regime, epoch length, and optimizer.

\paragraph{Results.} Figure~\ref{fig:experiments:optimizer} shows the results. As a baseline, RegNetY models show virtually no gap when trained using either SGD or AdamW (the difference \app0.1-0.2\% is within noise). On the other hand, \emph{\vit{P} models suffer a dramatic drop when trained with SGD} across all settings (of up to 10\% for larger models and longer training schedules). With a convolutional stem, \vit{C} models exhibit much smaller error gaps between SGD and AdamW across all training schedules and model complexities, including in larger models and longer schedules, where the gap is reduced to less than 0.2\%. In other words, both RegNetY and \vit{C} can be easily trained via either SGD or AdamW, but \vit{P} cannot.

\begin{figure}[t]\centering
\includegraphics[width=\linewidth]{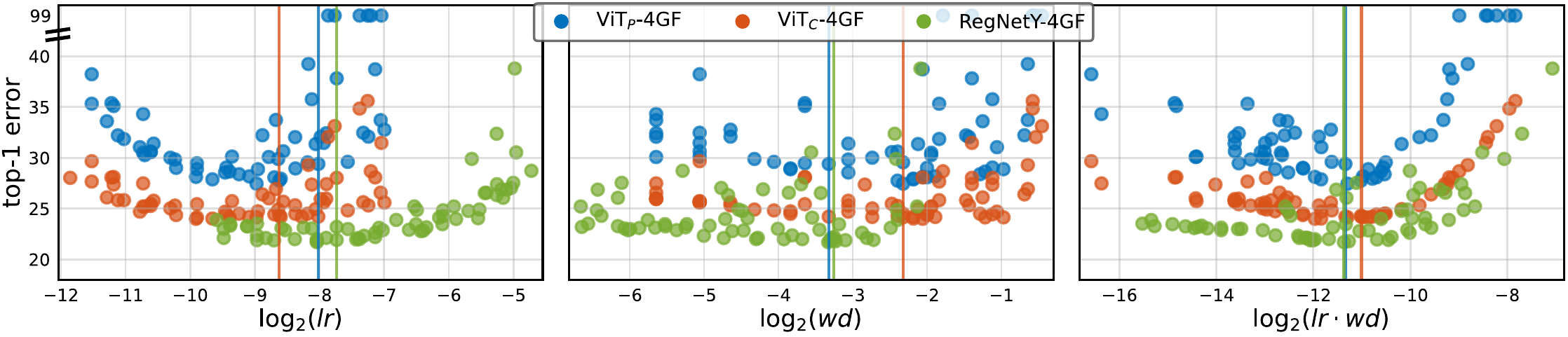}
\includegraphics[width=\linewidth]{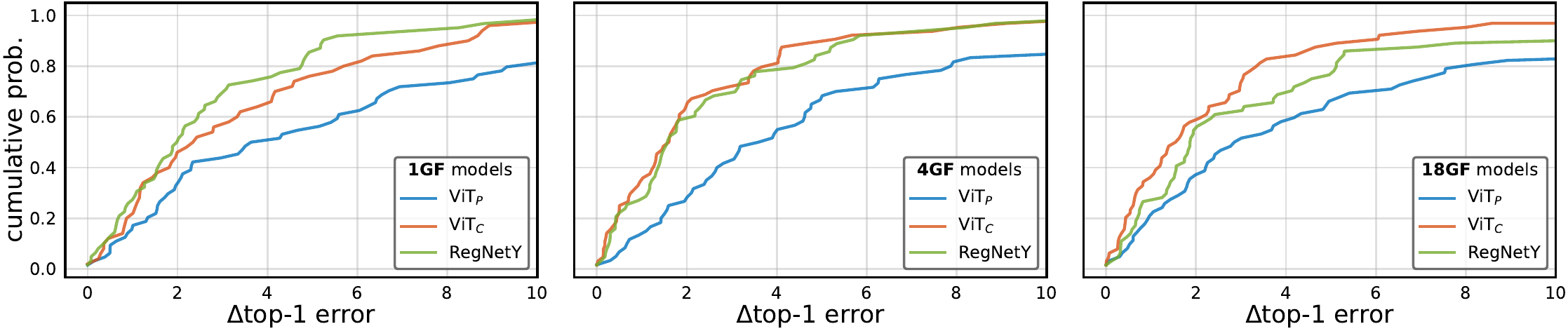}
\caption{\textbf{Hyperparameter stability for AdamW (\lr and \wtd}): For each model, we train 64 instances of the model for 50 epochs each with a random \lr and \wtd (in a fixed width interval around the optimal value for each model). \textit{Top}: Scatterplots of the \lr, \wtd, and \lr$\cdot$\wtd for three 4GF models. Vertical bars indicate optimal \lr, \wtd, and \lr$\cdot$\wtd values for each model. \textit{Bottom}: For each model, we generate an EDF of the errors by plotting the cumulative distribution of the $\Delta$top-1 errors ($\Delta$ to the optimal error for each model). A steeper EDF indicates better stability to \lr and \wtd variation. \vit{C} significantly improves the stability over the baseline \vit{P} across the model complexity spectrum, and matches or even outperforms the stability of the CNN model (RegNetY).}
\label{fig:lr_decay_adamw}
\end{figure}

\begin{figure}[t]\centering
\includegraphics[width=\linewidth]{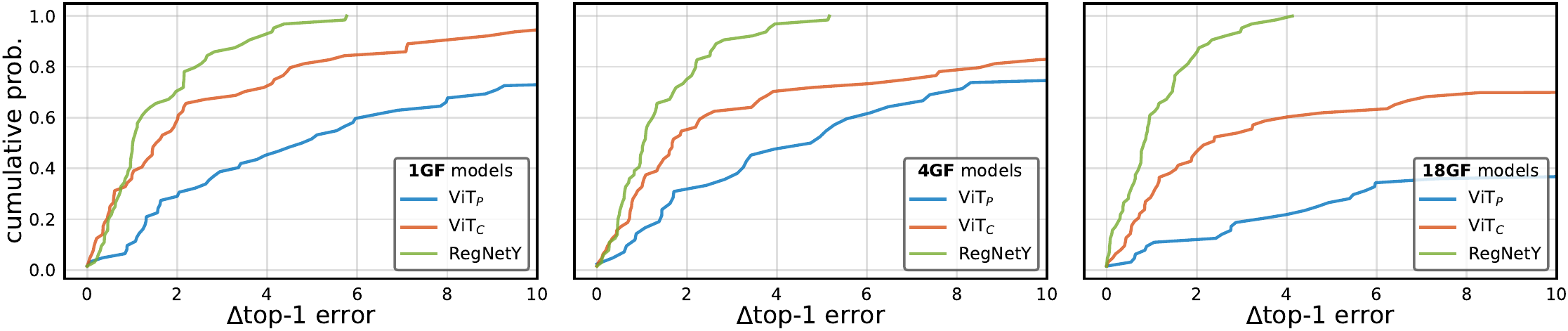}
\caption{\textbf{Hyperparameter stability for SGD (\lr and \wtd}): We repeat the setup from Figure~\ref{fig:lr_decay_adamw} using SGD instead of AdamW. The stability improvement of \vit{C} over the baseline \vit{P} is even larger than with AdamW. \Eg, \app60\% of \vit{C}{18GF} models are within 4\% $\Delta$top-1 error of the best result, while less than 20\% of \vit{P}{18GF} models are (in fact most \vit{P}{18GF} runs don't converge).}
\label{fig:lr_decay_sgd}
\end{figure}

\subsection{Learning Rate and Weight Decay Stability}\label{sec:experiments:lrwd}

Next, we characterize how sensitive different model families are to changes in learning rate (\lr) and weight decay (\wtd) under both AdamW and SGD optimizers. To quantify this, we make use of error distribution functions (EDFs)~\cite{Radosavovic2019}. An EDF is computed by sorting a set of results from low-to-high error and plotting the cumulative proportion of results as error increases, see~\cite{Radosavovic2019} for details. In particular, we generate EDFs of a model as a function of \lr and \wtd. The intuition is that if a model is robust to these hyperparameter choices, the EDF will be steep (all models will perform similarly), while if the model is sensitive, the EDF will be shallow (performance will be spread out).

We test 6 ViT models (\{$P$, $C$\} $\x$ \{1, 4, 18\} GF) and 3 RegNetY models (\{1, 4, 16\} GF). For each model and each optimizer, we compute an EDF by randomly sampling 64 (\lr, \wtd) pairs with learning rate and weight decay sampled in a fixed width interval around their optimal values for that model and optimizer (see the Appendix for sampling details). Rather than plotting absolute error in the EDF, we plot $\Delta$top-1 error between the best result (obtained with the optimal \lr and \wtd) and the observed result. Due to the large number of models, we train each for only 50 epochs.

\paragraph{Results.} Figure~\ref{fig:lr_decay_adamw} shows scatterplots and EDFs for models trained by AdamW. Figure~\ref{fig:lr_decay_sgd} shows SGD results. In all cases we see that \vit{C} significantly improves the \lr and \wtd stability over \vit{P} for both optimizers. This indicates that the \lr and \wtd are easier to optimize for \vit{C} than for \vit{P}.

\subsection{Experimental Details}\label{sec:experiments:details}

In all experiments we train with a single half-period cosine learning rate decay schedule with a 5-epoch linear learning rate warm-up~\cite{Goyal2017}. We use a minibatch size of $2048$. Crucially, weight decay is \emph{not} applied to the gain factors found in normalization layers nor to bias parameters anywhere in the model; we found that decaying these parameters can dramatically reduce top-1 accuracy for small models and short schedules. For inference, we use an exponential moving average (EMA) of the model weights (\eg,~\cite{dai2020fbnetv3}). The \lr and \wtd used in this section are reported in the Appendix. Other hyperparameters use defaults: SGD momentum is $0.9$ and AdamW's $\beta_1=0.9$ and $\beta_2=0.999$.

\paragraph{Regularization and data augmentation.} We use a simplified training recipe compared to recent work such as DeiT~\cite{Touvron2020training}, which we found to be equally effective across a wide spectrum of model complexities and dataset scales. We use AutoAugment~\cite{Cubuk2018}, mixup~\cite{Zhang2018mixup} ($\alpha=0.8$), CutMix~\cite{yun2019cutmix} ($\alpha=1.0$), and label smoothing~\cite{Szegedy2016a} ($\epsilon=0.1$). We prefer this setup because it is similar to common settings for CNNs (\eg, ~\cite{dollar2021fast}) except for stronger mixup and the addition of CutMix (ViTs benefit from both, while CNNs are not harmed). We compare this recipe to the one used for DeiT models in the Appendix, and observe that \emph{our setup provides substantially faster training convergence} likely because we remove repeating augmentation~\cite{berman2019multigrain,hoffer2019augment}, which is known to slow training~\cite{berman2019multigrain}.

\section{Peak Performance}\label{sec:reality}

A model's peak performance is the most commonly used metric in network design. It represents what is possible with the best-known-so-far settings and naturally evolves over time. Making fair comparisons between different models is desirable but fraught with difficulty. Simply citing results from prior work may be negatively biased against that work as it was unable to incorporate newer, yet applicable improvements. Here, we strive to provide a \emph{fairer comparison} between state-of-the-art CNNs, \vit{P}, and \vit{C}. We identify a set of factors and then strike a pragmatic balance between which subset to optimize for each model \vs which subset share a constant value across all models.

In our comparison, all models share the same epochs (400), use of model weight EMA, and set of regularization and augmentation methods (as specified in \S\ref{sec:experiments:details}). All CNNs are trained with SGD with \lr of $2.54$ and \wtd of $\expnumber{2.4}{-5}$; we found this single choice worked well across all models, as similarly observed in~\cite{dollar2021fast}. For all ViT models we found AdamW with a \lr/\wtd of $\expnumber{1.0}{-3}/0.24$ was effective, except for the 36GF models. For these larger models we tested a few settings and found a \lr/\wtd of $\expnumber{6.0}{-4}/0.28$ to be more effective for both \vit{P}{36GF} and \vit{C}{36GF} models. For training and inference, ViTs use $224\x 224$ resolution (we do \emph{not} fine-tune at higher resolutions), while the CNNs use (often larger) optimized resolutions specified in~\cite{dollar2021fast,Tan2019}. Given this protocol, we compare \vit{P}, \vit{C}, and CNNs across a spectrum of model complexities (1GF to 36GF) and dataset scales (directly training on ImageNet-1k \vs pretraining on ImageNet-21k and then fine-tuning on ImageNet-1k).

\begin{figure}[t]\centering
\includegraphics[width=0.99\linewidth]{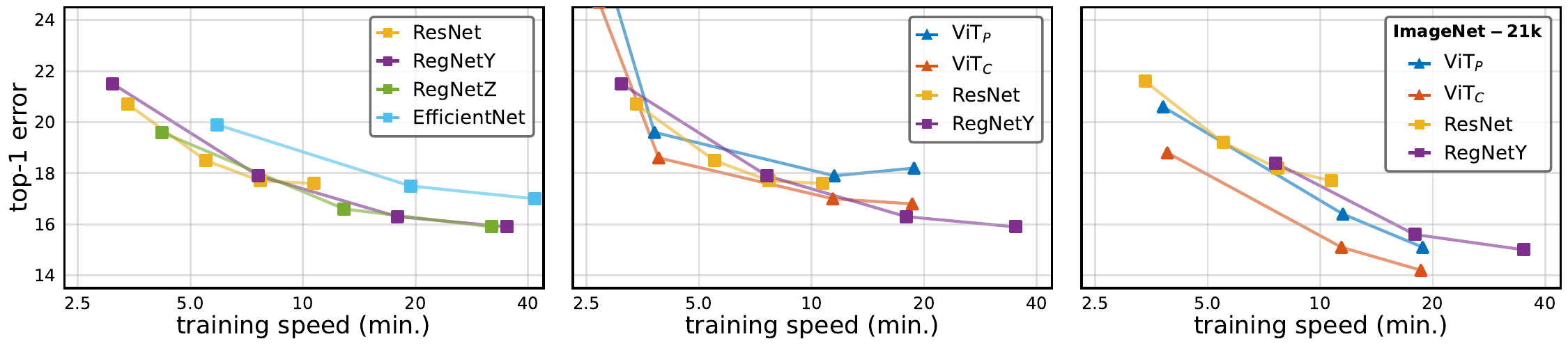}
\vspace{-2mm}
\caption{\textbf{Peak performance (epoch training time \vs ImageNet-1k val top-1 error)}: Results of a fair, controlled comparison of \vit{P}, \vit{C}, and CNNs. Each curve corresponds to a model complexity sweep resulting in a training speed spectrum (minutes per ImageNet-1k epoch). \emph{Left:} State-of-the-art CNNs. Equipped with a modern training recipe, ResNets are highly competitive in the faster regime, while RegNetY and Z perform similarly, and better than EfficientNets. \emph{Middle:} Selected CNNs compared to ViTs. With access to only ImageNet-1k training data, RegNetY \emph{and} ResNet outperform \vit{P} across the board. \vit{C} is more competitive with CNNs. \emph{Right:} Pretraining on ImageNet-21k improves the ViT models more than the CNNs, making \vit{P} competitive. Here, the proposed \vit{C} outperforms all other models across the full training speed spectrum.}
\label{fig:timeverror}
\end{figure}

\paragraph{Results.} Figure~\ref{fig:timeverror} shows a progression of results. Each plot shows ImageNet-1k val top-1 error \vs ImageNet-1k epoch training time.\footnote{We time models in PyTorch on 8 32GB Volta GPUs. We note that batch inference time is highly correlated with training time, but we report epoch time as it is easy to interpret and does not depend on the use case.} The left plot compares several state-of-the-art CNNs. RegNetY and RegNetZ~\cite{dollar2021fast} achieve similar results across the training speed spectrum and outperform EfficientNets~\cite{Tan2019}. Surprisingly, ResNets~\cite{He2016} are highly competitive at fast runtimes, showing that under a fairer comparison these years-old models perform substantially better than often reported (\cf~\cite{Tan2019}).

The middle plot compares two representative CNNs (ResNet and RegNetY) to ViTs, still using only ImageNet-1k training. The baseline \vit{P} underperforms RegNetY across the entire model complexity spectrum. To our surprise, \emph{\vit{P} also underperforms ResNets} in this regime. \vit{C} is more competitive and outperforms CNNs in the middle-complexity range.

\begin{table}[t]\centering\vspace{0mm}
\tablestyle{2pt}{1.05}\small
\resizebox{0.495\textwidth}{!}{\centering
\begin{tabular}{@{}l|ccccc|ccc|c@{}}
\multirow{2}{*}{model} & flops & params & acts & time & batch & \multicolumn{3}{c|}{epochs} & IN \\
 & (B) & (M) & (M) & (min) & size & ~100~ & ~200~& ~400~ & 21k \\\shline
ResNet-50    & \ph4.1 & \ph25.6 & 11.3   & \ph3.4    & 2048   & \gc{22.5} & \gc{21.2} & 20.7      & 21.6 \\
ResNet-101   & \ph7.8 & \ph44.5 & 16.4   & \ph5.5    & 2048   & \gc{20.3} & \gc{19.1} & 18.5      & 19.2 \\
ResNet-152   & 11.5   & \ph60.2 & 22.8   & \ph7.7    & 2048   & \gc{19.5} & \gc{18.4} & 17.7      & 18.2 \\
ResNet-200   & 15.0   & \ph64.7 & 32.3   & \bc{10.7} & 1024   & \gc{19.5} & \gc{18.3} & \bc{17.6} & \bc{17.7} \\\hline
RegNetY-1GF  & \ph1.0 & \phh9.6 & \ph6.2 & \ph3.1    & 2048   & \gc{23.2} & \gc{22.2} & 21.5      & - \\
RegNetY-4GF  & \ph4.1 & \ph22.4 & 14.5   & \ph7.6    & 2048   & \gc{19.4} & \gc{18.3} & 17.9      & 18.4 \\
RegNetY-16GF & 15.5   & \ph72.3 & 30.7   & \bc{17.9} & 1024   & \gc{17.1} & \gc{16.4} & \bc{16.3} & \bc{15.6} \\
RegNetY-32GF & 31.1   & 128.6   & 46.2   & 35.1      & \ph512 & \gc{16.2} & \gc{15.9} & 15.9      & 15.0 \\\hline
RegNetZ-1GF  & \ph1.0 & \ph11.0 & \ph8.8 & \ph4.2    & 2048   & \gc{20.8} & \gc{20.2} & 19.6      & - \\
RegNetZ-4GF  & \ph4.0 & \ph28.1 & 24.3   & \bc{12.9} & 1024   & \gc{17.4} & \gc{16.9} & \bc{16.6} & - \\
RegNetZ-16GF & 16.0   & \ph95.3 & 51.3   & 32.0      & \ph512 & \gc{16.0} & \gc{15.9} & 15.9      & - \\
RegNetZ-32GF & 32.0   & 175.1   & 79.6   & 55.3      & \ph256 & \gc{16.3} & \gc{16.2} & 16.1      & - \\
\end{tabular}}
\hfill
\resizebox{0.47\textwidth}{!}{\centering
\begin{tabular}{@{}l|ccccc|ccc|c@{}}
\multirow{2}{*}{model} & flops & params & acts & time & batch & \multicolumn{3}{c|}{epochs} & IN \\
& (B) & (M) & (M) & (min) & size & ~100~ & ~200~& ~400~ & 21k \\\shline
EffNet-B2 & \ph1.0 & \phh9.1 & 13.8   & \ph5.9    & 2048   & \gc{21.4} & \gc{20.5} & 19.9      & - \\
EffNet-B4 & \ph4.4 & \ph19.3 & 49.5   & \bc{19.4} & \ph512 & \gc{18.5} & \gc{17.8} & \bc{17.5} & - \\
EffNet-B5 & 10.3   & 30.4    & 98.9   & 41.7      & \ph256 & \gc{17.3} & \gc{17.0} & 17.0      & - \\
&&&& &&&  &\\\hline
\vit{P}{1GF}  & \ph1.1 & \phh4.8 & \ph5.5 & \ph2.6    & 2048   & \gc{33.2} & \gc{29.7} & 27.7      & - \\
\vit{P}{4GF}  & \ph3.9 & \ph18.5 & 11.1   & \ph3.8    & 2048   & \gc{23.3} & \gc{20.8} & 19.6      & 20.6 \\
\vit{P}{18GF} & 17.5   & \ph86.6 & 24.0   & 11.5      & 1024   & \gc{19.9} & \gc{18.4} & 17.9      & 16.4 \\
\vit{P}{36GF} & 35.9   & 178.4   & 37.3   & \bc{18.8} & \ph512 & \gc{19.9} & \gc{18.8} & \bc{18.2} & \bc{15.1}\\\hline
\vit{C}{1GF}  & \ph1.1 & \phh4.6 & \ph5.7 & \ph2.7    & 2048   & \gc{28.6} & \gc{26.1} & 24.7      & - \\
\vit{C}{4GF}  & \ph4.0 & \ph17.8 & 11.3   & \ph3.9    & 2048   & \gc{20.9} & \gc{19.2} & 18.6      & 18.8 \\
\vit{C}{18GF} & 17.7   & \ph81.6 & 24.1   & 11.4      & 1024   & \gc{18.4} & \gc{17.5} & 17.0      & 15.1 \\
\vit{C}{36GF} & 35.0   & 167.8   & 36.7   & \bc{18.6} & \ph512 & \gc{18.3} & \gc{17.6} & \bc{16.8} & \bc{14.2}\\
\end{tabular}}
\vspace{1.5mm}
\caption{\textbf{Peak performance (grouped by model family)}: Model complexity and validation top-1 error at 100, 200, and 400 epoch schedules on ImageNet-1k, and the top-1 error after pretraining on ImageNet-21k (IN 21k) and fine-tuning on ImageNet-1k. This table serves as reference for the results shown in Figure~\ref{fig:timeverror}. Blue numbers: best model trainable under 20 minutes per ImageNet-1k epoch. Batch sizes and training times are reported normalized to 8 32GB Volta GPUs (see Appendix). Additional results on the ImageNet-V2~\cite{recht2019imagenet} test set are presented in the Appendix.}
\label{tab:experiments:reality}
\vspace{-5mm}
\end{table}

The right plot compares the same models but with ImageNet-21k pretraining (details in Appendix). In this setting ViT models demonstrates a greater capacity to benefit from the larger-scale data: now \vit{C} strictly outperforms both \vit{P} and RegNetY. Interestingly, \emph{the original \vit{P} does not outperform a state-of-the-art CNN} even when trained on this much larger dataset. Numerical results are presented in Table~\ref{tab:experiments:reality} for reference to exact values. This table also highlights that flop counts are not significantly correlated with runtime, but that activations are (see Appendix for more details), as also observed by~\cite{dollar2021fast}. \Eg, EfficientNets are slow relative to their flops while ViTs are fast.

These results verify that \vit{C}'s convolutional stem improves not only optimization stability, as seen in the previous section, but also peak performance. Moreover, this benefit can be seen across the model complexity and dataset scale spectrum. Perhaps surprisingly, given the recent excitement over ViT, we find that \vit{P} struggles to compete with state-of-the-art CNNs. We only observe improvements over CNNs when using \emph{both} large-scale pretraining data \emph{and} the proposed convolutional stem.

\section{Conclusion}\label{sec:conclusion}

In this work we demonstrated that the optimization challenges of ViT models are linked to the large-stride, large-kernel convolution in ViT's patchify stem. The seemingly trivial change of replacing this patchify stem with a simple convolutional stem leads to a remarkable change in optimization behavior. With the convolutional stem, ViT (termed \vit{C}) converges faster than the original ViT (termed \vit{P}) (\S\ref{sec:experiments:length}), trains well with either AdamW or SGD (\S\ref{sec:experiments:optimizer}), improves learning rate and weight decay stability (\S\ref{sec:experiments:lrwd}), and improves ImageNet top-1 error by \app 1-2\% (\S\ref{sec:reality}). These results are consistent across a wide spectrum of model complexities (1GF to 36GF) and dataset scales (ImageNet-1k to ImageNet-21k). Our results indicate that injecting a small dose of convolutional inductive bias into the early stages of ViTs can be hugely beneficial. Looking forward, we are interested in the theoretical foundation of why such a minimal architectural modification can have such large (positive) impact on optimizability. We are also interested in studying larger models. Our preliminary explorations into 72GF models reveal that the convolutional stem still improves top-1 error, however we also find that a \emph{new} form of instability arises that causes training error to randomly spike, especially for \vit{C}.

{\small \paragraph{Acknowledgements.} We thank Herv\'e Jegou, Hugo Touvron, and Kaiming He for valuable feedback.}

\clearpage\appendix
\section*{Appendix A: Stem Design Ablation Experiments}

ViT's patchify stem differs from the proposed convolutional stem in the type of convolution used and the use of normalization and a non-linear activation function. We investigate these factors next.

\begin{table*}
\tablestyle{3.0pt}{1.05}
\resizebox{1.0\textwidth}{!}{\centering
\begin{tabular}{@{}lllll|ccc|ccc@{}}
\multirow{2}{*}{stem} & \multirow{2}{*}{kernel size} & \multirow{2}{*}{stride} & \multirow{2}{*}{padding} & \multirow{2}{*}{channels} & flops & params & acts & \multicolumn{2}{c}{top-1 error} & \multirow{2}{*}{$\Delta$} \\
 & &&&& (M) & (M) & (M) & AdamW & SGD \\\shline
$P$ & [\bc{16}] & [\bc{16}] & [0] & [384] & 58 & 0.3 & 0.8 & 27.7 & 33.0 & 5.3 \\
$C$ & [3, 3, 3, 3, 1] & [2, 2, 2, 2, 1] & [1, 1, 1, 1, 0] & [48, 96, 192, 384, 384] & 435 & 1.0 & 1.2 & \bf{24.0} & \bf{24.7} & \bf{0.7} \\\hline
$S1$ & [3, 3, 3, \bc{2}, 1] & [2, 2, 2, \bc{2}, 1] & [1, 1, 1, 0, 0] & [42, 104, 208, 416, 384] & 422 & 0.8 & 1.3 & 24.3 & 25.1 & 0.8 \\
$S2$ & [3, 3, 3, \bc{4}, 1] & [2, 2, 1, \bc{4}, 1] & [1, 1, 1, 0, 0] & [32, 64, 128, 256, 384] & 422 & 0.7 & 1.1 & 24.3 & 25.3 & 1.0 \\
$S3$ & [3, 3, 3, \bc{8}, 1] & [2, 1, 1, \bc{8}, 1] & [1, 1, 1, 0, 0] & [17, 34, 68, 136, 384] & 458 & 0.7 & 1.6 & 25.1 & 26.2 & 1.1 \\
$S4$ & [3, 3, 3, \bc{16}, 1] & [1, 1, 1, \bc{16}, 1] & [1, 1, 1, 0, 0] & [8, 16, 32, 64, 384] & 407 & 0.6 & 2.9 & 26.2 & 27.9 & 1.3 \\
\end{tabular}}
\caption{\textbf{Stem designs}: We compare ViT's standard patchify stem ($P$) and our convolutional stem ($C$) to four alternatives ($S1$ - $S4$) that each include a \emph{patchify layer}, \ie, a convolution with kernel size ($> 1$) equal to stride (highlighted in blue). Results use 50 epoch training, 4GF model size, and optimal \lr and \wtd values for all models. We observe that increasing the pixel size of the patchify layer ($S1$ - $S4$) systematically degrades both top-1 error and optimizer stability ($\Delta$) relative to $C$.}
\label{tab:ablation:stem_design}
\end{table*}

\begin{figure}[t]\centering
\includegraphics[width=\linewidth]{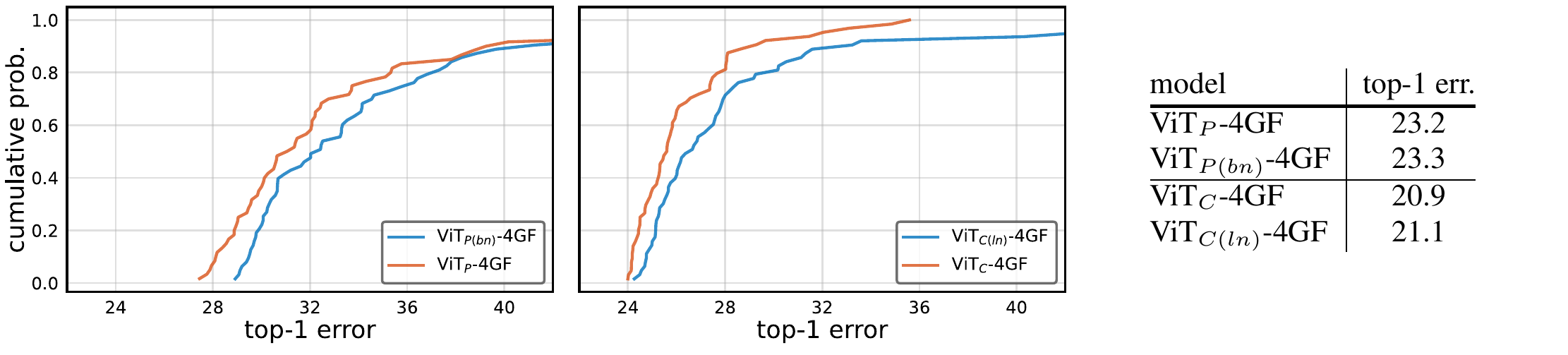}
\caption{\textbf{Stem normalization and non-linearity}: We apply BN and ReLU after the patchify stem and train \vit{P}{4GF} (\emph{left plot}), or replace BN with layer norm (LN) in the convolutional stem of \vit{C}{4GF} (\emph{middle plot}). EDFs are computed by sampling \lr and \wtd values and training for 50 epochs. The table (\emph{right}) shows 100 epoch results using best \lr and \wtd values found at 50 epochs. The minor gap in error in the EDFs and at 100 epochs indicates that these choices are fairly insignificant.}
\label{fig:bn_ln_ablation}
\end{figure}

\paragraph{Stem design.} The focus of this paper is studying the large, positive impact of changing ViT's default patchify stem to a simple, standard convolutional stem constructed from stacked stride-two $3\x 3$ convolutions. Exploring the stem design space, and more broadly ``hybrid ViT'' models~\cite{Dosovitskiy2020image}, to maximize peak performance is an explicit \emph{anti-goal} because we want to study the impact under minimal modifications. However, we can gain additional insight by considering alternative stem designs that fall between the patchify stem ($P$) the standard convolutional stem ($C$). Four alternative designs ($S1$ - $S4$) are presented in Table~\ref{tab:ablation:stem_design}. The stems are designed so that overall model flops remain comparable. Stem $S1$ modifies $C$ to include a small $2\x 2$ patchify layer, which slightly worsens results. Stems \mbox{$S2$ - $S4$} systematically increase the pixel size $p$ of the patchify layer from $p=2$ up to $16$, matching the size used in stem $P$. \emph{Increasing $p$ reliably degrades both error and optimizer stability.} Although we selected the $C$ design \emph{a priori} based on existing best-practices for CNNs, we see \emph{ex post facto} that it outperforms four alternative designs that each include one patchify layer.

\paragraph{Stem normalization and non-linearity.} We investigate normalization and non-linearity from two directions: (1) adding BN and ReLU to the default patchify stem of ViT, and (2) changing the normalization in the proposed convolutional stem. In the first case, we simply apply BN and ReLU after the patchify stem and train \vit{P}{4GF} (termed \vit{P(bn)}{4GF}) for 50 and 100 epochs. For the second case, we run four experiments with \vit{C}{4GF}: \{50, 100\} epochs $\x$ \{BN, layer norm (LN)\}. As before, we tune \lr and \wtd for each experiment using the 50-epoch schedule and reuse those values for the 100-epoch schedule. We use AdamW for all experiments. Figure~\ref{fig:bn_ln_ablation} shows the results. From the EDFs, which use a 50 epoch schedule, we see that the addition of BN and ReLU to the patchify stem slightly worsens the best top-1 error but does not affect \lr and \wtd stability (\emph{left}). Replacing BN with LN in the convolutional stem marginally degrades both best top-1 error and stability (\emph{middle}). The table (\emph{right}) shows 100 epoch results using optimal \lr and \wtd values chosen from the 50 epoch runs. At 100 epochs the error gap is small indicating that these factors are likely insignificant.

\begin{figure}\centering
\includegraphics[width=\linewidth]{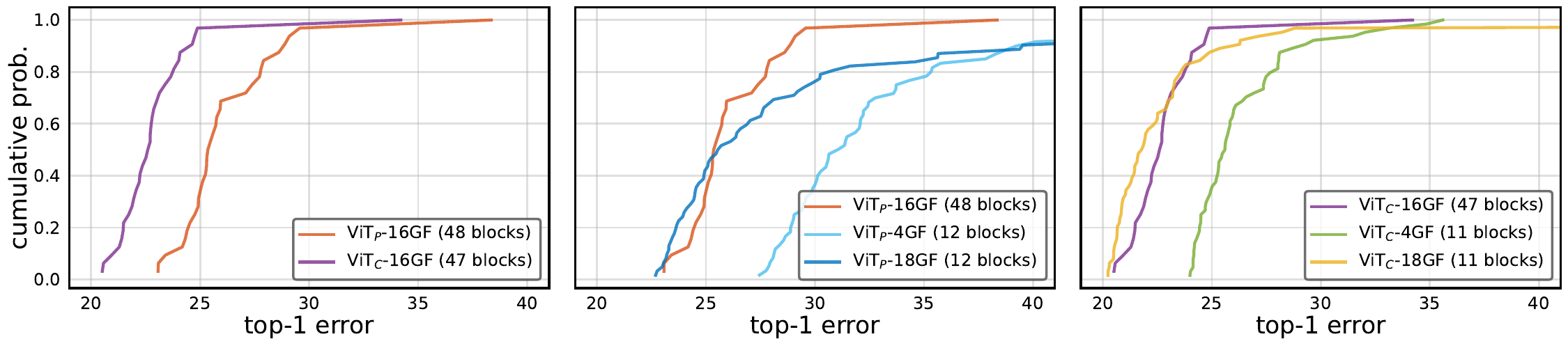}
\caption{\textbf{Deeper models}: We increase the depth of \vit{P}{4GF} from 12 to 48 blocks, termed as \vit{P}-16GF (48 blocks), and create a counterpart with a convolutional stem, \vit{C}-16GF (47 blocks); all models are trained for 50 epochs. \emph{Left}: The convolutional stem significantly improves error and stability despite accounting for only $\app2\%$ total flops. \emph{Middle}, \emph{Right}: The deeper 16GF ViTs clearly outperform the shallower 4GF models and achieve similar (slightly worse) error to the shallower and wider 18GF models. The deeper \vit{P} also has better \lr/\wtd stability than the shallower \vit{P} models.}
\label{fig:appx:lr_decay_deepmodel_adamw}
\end{figure}

\begin{figure}[t]\centering
\includegraphics[width=\linewidth]{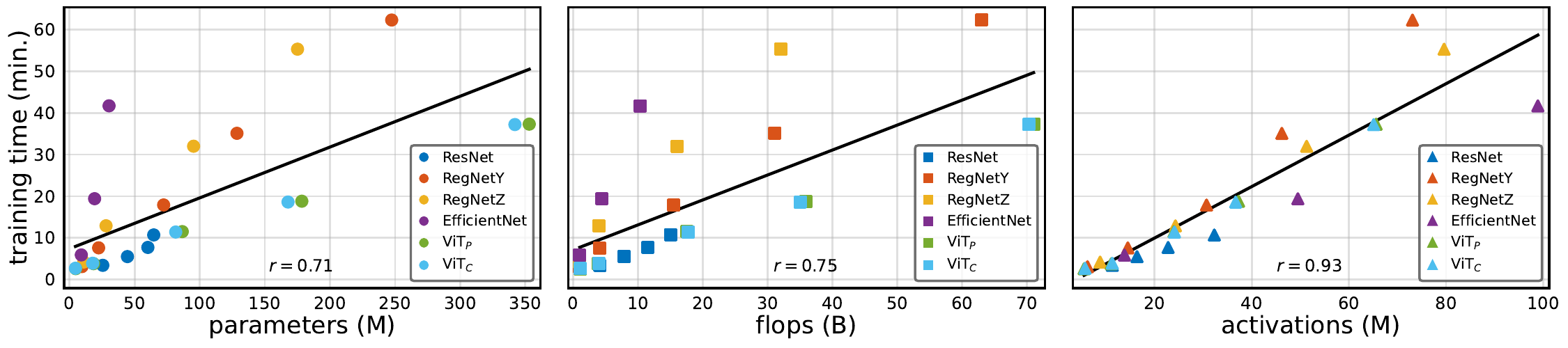}
\caption{\textbf{Complexity measures \vs runtime}: We plot the GPU runtime of models versus three commonly used complexity measures: \emph{parameters}, \emph{flops}, and \emph{activations}. For all models, including ViT, \emph{runtime is most correlated with activations}, not flops, as was previously shown for CNNs~\cite{dollar2021fast}.}
\label{fig:appx:complexity_measures_vs_time}\vspace{-2mm}
\end{figure}

\section*{Appendix B: Deeper Model Ablation Experiments}

Touvron \etal~\cite{touvron2021going} found that deeper ViT models are more unstable, \eg, increasing the number of transformer blocks from 12 to 36 may cause a \app10 point drop in top-1 accuracy given a fixed choice of \lr and \wtd. They demonstrate that stochastic depth and/or their proposed LayerScale can remedy this training failure. Here, we explore deeper models by looking at EDFs created by sampling \lr and \wtd. We increase the depth of a \vit{P}{4GF} model from 12 blocks to 48 blocks, termed \vit{P}-16GF (48 blocks). We then remove one block and use the convolutional stem from \vit{C}{4GF}, yielding a counterpart \vit{C}-16GF (47 blocks) model. Figure~\ref{fig:appx:lr_decay_deepmodel_adamw} shows the EDFs of the two models and shallower models for comparison, following the setup in \S\ref{sec:experiments:lrwd}. Despite the convolutional stem accounting for only $1/48$ ($\app2\%$) total flops, it shows solid improvement over its patchify counterpart. We find that a variety of \lr and \wtd choices allow deeper ViT models to be trained without a large drop in top-1 performance and without additional modifications. In fact, the deeper \vit{P}{16GF} (48 blocks) has better \lr and \wtd stability than \vit{P}{4GF} and \vit{P}{18GF} over the sampling range (Figure~\ref{fig:appx:lr_decay_deepmodel_adamw}, \emph{middle}).

\section*{Appendix C: Larger Model ImageNet-21k Experiments}
In Table~\ref{tab:experiments:reality} we reported the peak performance of ViT models on ImageNet-21k up to 36GF. To study larger models, we construct a 72GF \vit{P} by using 22 blocks, 1152 hidden size, 18 heads, and 4 MLP multiplier. For \vit{C}{72GF}, we use the same $C$-stem design used for \vit{C}{18GF} and \vit{C}{36GF}, but \emph{without} removing one transformer block since the flops increase from the $C$-stem is marginal in this complexity regime.

Our preliminary explorations into 72GF ViT models directly adopted hyperparameters used for 36GF ViT models. Under this setting, we observed that the convolutional stem still improves top-1 error, however, we also found that a new form of instability arises, which causes training error to randomly spike. Sometimes training may recover within the same epoch, and subsequently the final accuracy is not impacted; or, it may take several epochs to recover from the error spike, and in this case we observe suboptimal final accuracy. The first type of error spike is more common for \vit{P}{72GF}, while the latter type of error spike is more common for \vit{C}{72GF}.

To mitigate this instability, we adopt two measures: (i) For both models, we lower \wtd from $0.28$ to $0.15$ as we found that it significantly reduces the chance of error spikes. (ii) For \vit{C}{72GF}, we initialize its stem from the ImageNet-21k pre-trained \vit{C}{36GF} and keep it frozen throughout training. These modifications make training \vit{}{72GF} models on ImageNet-21k feasible. When fine-tuned on ImageNet-1k, \vit{P}{72GF} reaches $14.2\%$ top-1 error and \vit{C}{72GF} reaches $13.6\%$ top-1 error, showing that \vit{C} still outperforms its \vit{P} counterpart. Increasing fine-tuning resolution from $224$ to $384$ boosts the performance of \vit{C}{72GF} to $12.6\%$ top-1 error, while significantly increasing the fine-tuning model complexity from 72GF to 224GF.

\section*{Appendix D: Model Complexity and Runtime}

In previous sections, we reported error \vs training time. Other commonly used complexity measures include \emph{parameters}, \emph{flops}, and \emph{activations}. Indeed, it is most typical to report accuracy as a function of model flops or parameters. However, flops may fail to reflect the bottleneck on modern memory-bandwidth limited accelerators (\eg, GPUs, TPUs). Likewise, parameters are an even more unreliable predictor of model runtime. Instead, activations have recently been shown to be a better proxy of runtime on GPUs (see~\cite{Radosavovic2020,dollar2021fast}). We next explore if similar results hold for ViT models.

For CNNs, previous studies~\cite{Radosavovic2020,dollar2021fast} defined \emph{activations} as the \emph{total size of all output tensors of the convolutional layers}, while disregarding normalization and non-linear layers (which are typically paired with convolutions and would only change the activation count by a constant factor). In this spirit, for transformers, we define \emph{activations as the size of output tensors of all matrix multiplications}, and likewise disregard element-wise layers and normalizations. For models that use both types of operations, we simply measure the output size of all convolutional and vision transformer layers.

Figure~\ref{fig:appx:complexity_measures_vs_time} shows the runtime as a function of these model complexity measures. The Pearson correlation coefficient ($r$) confirms that activations have a much stronger linear correlation with actual runtime ($r=0.93$) than flops ($r=0.75$) or parameters ($r=0.71$), confirming that the findings of~\cite{dollar2021fast} for CNNs also apply to ViTs. While flops are somewhat predictive of runtime, models with a large ratio of activations to flops, such as EfficientNet, have much higher runtime than expected based on flops. Finally, we note that \vit{P} and \vit{C} are nearly identical on all complexity measures and runtime.

\paragraph{Timing.} Throughout the paper we report \emph{normalized} training time, as if the model were trained on a single 8 V100 GPU server, by multiplying the actual training time by the number of GPUs used and dividing by 8. (Due to different memory requirements of different models, we may be required to scale up the number of GPUs to accommodate the target minibatch size.) We use the number of minutes taken to process one ImageNet-1k epoch as a standard unit of measure. We prefer training time over inference time because inference time depends heavily on the use case (\eg, a streaming, latency-oriented setting requires a batch size of $1$ \vs a throughput-oriented setting that allows for batch size $\gg 1$) and the hardware platform (\eg, smartphone, accelerator, server CPU).

\begin{table}[t]
\tablestyle{8.0pt}{1.05}
\resizebox{0.4\textwidth}{!}{\centering
\begin{tabular}{@{}l|cc|cc@{}}
\multirow{2}{*}{model} & \multicolumn{2}{c|}{AdamW} & \multicolumn{2}{c}{SGD} \\
& \lr & \wtd & \lr & \wtd \\\shline
RegNetY-$\ast$ & 3.8e-3 & 0.1 & 2.54 & 2.4e-5 \\\hline
\vit{P}{1GF} & 2.0e-3 & 0.20 & 1.9 & 1.3e-5 \\
\vit{P}{4GF} & 2.0e-3 & 0.20 & 1.9 & 1.3e-5 \\
\vit{P}{18GF} & 1.0e-3 & 0.24 & 1.1 & 1.2e-5 \\\hline
\vit{C}{1GF} & 2.5e-3 & 0.19 & 1.9 & 1.3e-5 \\
\vit{C}{4GF} & 1.0e-3 & 0.24 & 1.3 & 2.2e-5 \\
\vit{C}{18GF} & 1.0e-3 & 0.24 & 1.1 & 2.7e-5 \\
\end{tabular}}
\hspace{10mm}
\resizebox{0.45\textwidth}{!}{\centering
\begin{tabular}{@{}l|cc@{}}
\multirow{2}{*}{model} & \multicolumn{2}{c}{AdamW} \\
& \lr & \wtd \\\shline
\vit{}-$\ast$ & $(\expnumber{2.5}{-4}, \expnumber{8.0}{-3})$ & $(0.02, 0.8)$ \\
RegNetY-$\ast$ & $(\expnumber{1.25}{-3}, \expnumber{4.0}{-2})$ & $(0.0075, 0.24)$ \\
\multicolumn{3}{c}{~}\\
\multirow{2}{*}{model} & \multicolumn{2}{c}{SGD} \\
& \lr & \wtd \\\shline
\vit{}-$\ast$ & $(0.1, 3.2)$ & $(\expnumber{4.0}{-6}, \expnumber{1.2}{-4})$ \\
RegNetY-$\ast$ & $(0.25, 8.0)$ & $(\expnumber{3.0}{-6}, \expnumber{8.0}{-5})$ \\
\end{tabular}}
\vspace{1.5mm}
\caption{\textbf{Learning rate and weight decay used in \S\ref{sec:experiments}}: \emph{Left}: Per-model \lr and \wtd values used for the experiments in \S\ref{sec:experiments:length} and \S\ref{sec:experiments:optimizer}, optimized for ImageNet-1k at 50 epochs. \emph{Right}: Per-model \lr and \wtd ranges used for the experiments in \S\ref{sec:experiments:lrwd}. Note that for our final experiments in \S\ref{sec:reality}, we constrained the \lr and \wtd values further, using a single setting for all CNN models, and just two settings for all ViT models. We recommend using this simplified set of values in \S\ref{sec:reality} when comparing models for fair and easily reproducible comparisons. All \lr{} values are normalized \wrt a minibatch size of $2048$~\cite{Goyal2017}.}
\label{tab:details:optimal_lr_wd}\vspace{-4mm}
\end{table}

\section*{Appendix E: Additional Experimental Details}

\paragraph{Stability experiments.} For the experiments in \S\ref{sec:experiments:length} and \S\ref{sec:experiments:optimizer}, we allow each CNN and ViT model to select a different \lr and \wtd. We find that all CNNs select nearly identical values, so we normalize them to a single choice as done in~\cite{dollar2021fast}. ViT models prefer somewhat more varied choices. Table~\ref{tab:details:optimal_lr_wd} (\emph{left}) lists the selected values. For the experiments in \S\ref{sec:experiments:lrwd}, we use \lr and \wtd intervals shown in Table~\ref{tab:details:optimal_lr_wd} (\emph{right}). These ranges are constructed by (i) obtaining initial good \lr and \wtd choices for each model family; and then (ii) multiplying them by $1/8$ and $4.0$ for left and right interval endpoints (we use an asymmetric interval because models are trainable with smaller but not larger values). Finally we note that if we were to redo the experiments, the setting used in \S\ref{sec:experiments:length}/\S\ref{sec:experiments:optimizer} could be simplified.

\paragraph{Peak performance on ImageNet-1k.}  We note that in later experiments we found tuning \lr and \wtd per model is \emph{not} necessary to obtain competitive results. Therefore, for our final experiments in \S\ref{sec:reality}, we constrained the \lr and \wtd values further, using a single setting for all CNN models, and just two settings for all ViT models, as discussed in \S\ref{sec:reality}. We recommend using this simplified set of values when comparing models for fair and easily reproducible comparisons. Finally, for these experiments, when training is memory constrained (\ie, for EfficientNet-\{B4,B5\}, RegNetZ-\{4,16,32\}GF), we reduce the minibatch size from $2048$ and linearly scale the \lr according to~\cite{Goyal2017}.

\paragraph{Peak performance on ImageNet-21k.} For ImageNet-21k, a dataset of 14M images and \app 21k classes, we pretrain models for 90 (ImageNet-21k) epochs, following~\cite{Dosovitskiy2020image}. We do \emph{not} search for the optimal settings for ImageNet-21k and instead use the identical training recipe (up to minibatch size) used for ImageNet-1k. To reduce training time, we distribute training over more GPUs and use a larger minibatch size of $4096$ with the \lr scaled accordingly. For simplicity and reproducibility, we use a single label per image, unlike some prior work (\eg,~\cite{tan2021efficientnetv2,ridnik2021imagenet}) that uses WordNet~\cite{miller1995wordnet} to expand single labels to multiple labels. After pretraining, we fine-tune for 20 epochs on ImageNet-1k and use a small-scale grid search of \lr while  keeping \wtd at 0, similar to~\cite{Dosovitskiy2020image,tan2021efficientnetv2}.

\begin{table*}[t]
\tablestyle{8.0pt}{1.1}
\resizebox{0.6\textwidth}{!}{\centering
\begin{tabular}{@{}c|cccccccc|ccc@{}}
model
& \rotatebox{90}{Augment} & \rotatebox{90}{Mixup} & \rotatebox{90}{CutMix}
& \rotatebox{90}{Label Smooth} & \rotatebox{90}{Model EMA} & \rotatebox{90}{Erasing}
& \rotatebox{90}{Stoch Depth} & \rotatebox{90}{Repeating}
& \rotatebox{90}{100 epochs} & \rotatebox{90}{400 epochs}
& \rotatebox{90}{300 epochs~\cite{Touvron2020training}} \\\shline
\multirow{6}{*}{\rotatebox{0}{\vit{P}{4GF}}}
& Auto & \cmark & \cmark & \cmark & \cmark & & & & 23.2 & 20.5 & \gc{-} \\
& Rand & \cmark & \cmark & \cmark & & \cmark & \cmark & \cmark & 25.4 & 20.7 & \gc{-} \\
& Rand & \cmark & \cmark & \cmark & & \cmark & & \cmark & 24.9 & 20.5 & \gc{-}\\
& Rand & \cmark & \cmark & \cmark & & \cmark & & & 23.6 & 20.4 & \gc{-} \\
& Rand & \cmark & \cmark & \cmark & & & & & 23.5 & 20.3 & \gc{-} \\
& Auto & \cmark & \cmark & \cmark & & & & & 23.0 & 20.3 & \gc{-} \\
\hline
\multirow{9}{*}{\rotatebox{0}{\vit{P}{18GF}}}
& Auto & \cmark & \cmark & \cmark & \cmark & & & & 19.9 & 17.9 & \gc{-} \\
& Rand & \cmark & \cmark & \cmark & & \cmark & \cmark & \cmark & 22.5 & 18.6 & \gc{18.2} \\
& Rand & \cmark & \cmark & \cmark & & \cmark & & \cmark & 25.1 & 19.2 & \gc{96.6} \\
& Rand & \cmark & \cmark & \cmark & & \cmark & & & 21.2 & 19.9 & \gc{-} \\
& Rand & \cmark & \cmark & \cmark & & & & & 20.9 & 19.7 & \gc{-} \\
& Auto & \cmark & \cmark & \cmark & & & & & 20.4 & 20.0 & \gc{-} \\
& Rand & \cmark & \cmark & \cmark & & \cmark & \cmark & & - & - & \gc{22.6} \\
& Rand & \cmark & \cmark & \cmark & & & \cmark & \cmark & - & - & \gc{95.7} \\
& Rand & \cmark & \cmark & \cmark & \cmark & \cmark & \cmark & \cmark & - & - & \gc{18.1} \\
\end{tabular}}
\caption{\textbf{Ablation of data augmentation and regularization}: We use the \lr and \wtd from Table~\ref{tab:details:optimal_lr_wd} (left), except for \vit{P}{18GF} models with RandAugment which benefit from stronger \wtd (we increase \wtd to 0.5). Original DeiT ablation results~\cite{Touvron2020training} are copied for reference in gray (\emph{last column}); these use a \lr/\wtd of $\expnumber{1}{-3}/0.05$ (\lr normalized to minibatch size $2048$), which leads to some training failures (we note our \wtd is 5-10$\x$ higher). Our default training setup (\emph{first row} in each set) uses AutoAugment, mixup, CutMix, label smoothing, and model EMA. Compared to the DeiT setup (\emph{second row} in each set), we do not use erasing, stochastic depth, or repeating. Although our setup is equally effective, it is simpler and also converges much faster (see Figure~\ref{fig:appx:training_length_deit}).}
\label{tab:ablation:augmentations}
\end{table*}

\begin{figure}[t]\centering
\includegraphics[width=\linewidth]{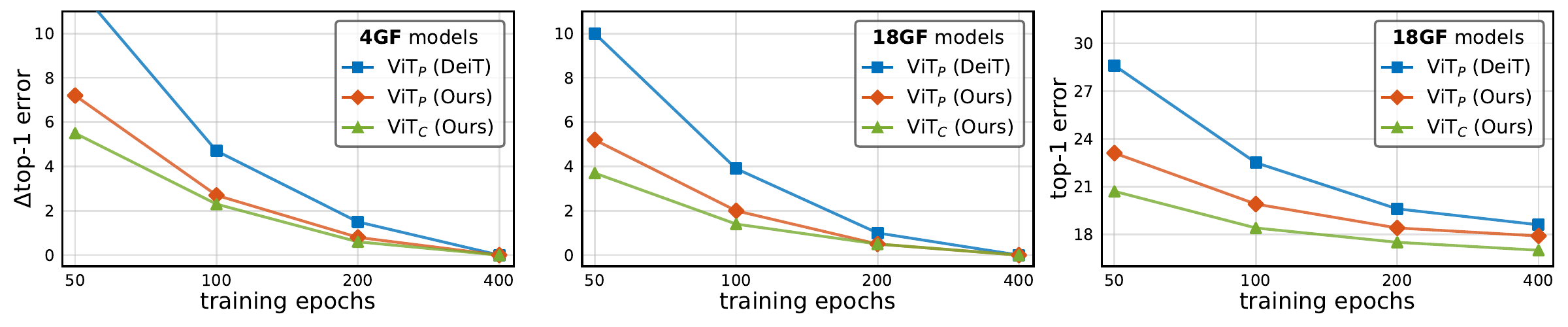}
\caption{\textbf{Impact of training recipes on convergence}: We train ViT models using the DeiT recipe \vs our simplified counterpart. \emph{Left and middle}: $\Delta$top-1 error of 4GF and 18GF models at 50, 100 and 200 epoch schedules, and asymptotic performance at 400 epochs. \emph{Right}: Absolute top-1 error of 18GF models. Removing augmentations and using model EMA accelerates convergence for both \vit{P} and \vit{C} models while slightly improving upon our reproduction of DeiT's top-1 error.}
\label{fig:appx:training_length_deit}
\end{figure}

\section*{Appendix F: Regularization and Data Augmentation}

At this study's outset, we developed a simplified training setup for ViT models. Our goals were to design a training setup that is as simple as possible, resembles the setup used for state-of-the-art CNNs~\cite{dollar2021fast}, and maintains competitive accuracy with DeiT~\cite{Touvron2020training}. Here, we document this exploration by considering the baseline \vit{P}{4GF} and \vit{P}{18GF} models. Beyond simplification, we also observe that our training setup yields faster convergence than the DeiT setup, as discussed below.

Table~\ref{tab:ablation:augmentations} compares our setup to that of DeiT~\cite{Touvron2020training}. Under their \lr/\wtd choice, \cite{Touvron2020training} report failed training when removing~\emph{erasing} and~\emph{stochastic depth}, as well as significant drop of accuracy when removing~\emph{repeating}. We find that they can be safely disabled as long as a higher \wtd is used (our \wtd is 5-10$\x$ higher). We observe that we can remove model EMA for \vit{P}{4GF}, but that it is essential for the larger \vit{P}{18GF} model, especially at 400 epochs. Without model EMA, \vit{P}{18GF} can still be trained effectively, but this requires additional augmentation and regularization (as in DeiT).

Figure~\ref{fig:appx:training_length_deit} shows that our training setup accelerates convergence for both \vit{P} and \vit{C} models, as can be seen by comparing the error \emph{deltas} ($\Delta$top-1) between the DeiT baseline and ours (\emph{left} and \emph{middle} plots). Our training setup also yields slightly better top-1 error than our reproduction of DeiT (\emph{right} plot). We conjecture that faster convergence is due to removing repeating augmentation~\cite{berman2019multigrain,hoffer2019augment}, which was shown in~\cite{berman2019multigrain} to slow convergence. Under some conditions repeating augmentation may improve accuracy, however we did not observe such improvements in our experiments.

\begin{figure}[t]\centering
\includegraphics[width=0.42\linewidth]{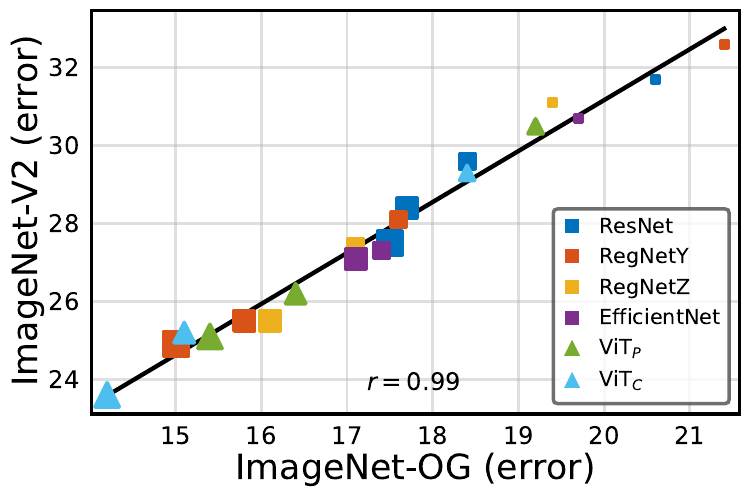}
\caption{\textbf{ImageNet-V2 performance}: We take the models from Table~\ref{tab:experiments:reality} and benchmark them on the ImageNet-V2 test set. Top-1 errors are plotted for the original (OG) ImageNet validation set (x-axis) and the ImageNet-V2 test set (y-axis). Rankings are mostly preserved up to one standard deviation of noise (estimated at $\app 0.1$-$0.2\%$) and the two testing sets exhibit linear correlation (Pearson's $r=0.99$). Marker size corresponds to model flops.}
\label{fig:appx:inv2}
\end{figure}

\section*{Appendix G: ImageNet-V2 Evaluation}
In the main paper and previous appendix sections we benchmarked all models on the original (OG) ImageNet validation set~\cite{Deng2009}. Here we benchmark our models on the ImageNet-V2~\cite{recht2019imagenet}, a new test
set collected following the original procedure. We take the 400-epoch \emph{or} ImageNet-21k models from Table~\ref{tab:experiments:reality}, depending on which one is better, and evaluate them on ImageNet-V2 to collect top-1 errors. Figure~\ref{fig:appx:inv2} shows that rankings are mostly preserved up to one standard deviation of noise (estimated at $\app 0.1$-$0.2\%$). The two testing sets exhibit linear correlation, as confirmed by the Pearson correlation coefficient $r = 0.99$, despite ImageNet-V2 results showing higher absolute error. The parameters of the fit line are given by $y = 1.31x + 5.0$.

{\setlength{\bibsep}{2.9pt}\small\bibliographystyle{plain}\bibliography{vitconv}}

\begin{thebibliography}{10}

\bibitem{berman2019multigrain}
Maxim Berman, Herv{\'e} J{\'e}gou, Andrea Vedaldi, Iasonas Kokkinos, and
  Matthijs Douze.
\newblock {MultiGrain}: a unified image embedding for classes and instances.
\newblock {\em arXiv:1902.05509}, 2019.

\bibitem{Buades2005non}
Antoni Buades, Bartomeu Coll, and J-M Morel.
\newblock A non-local algorithm for image denoising.
\newblock In {\em CVPR}, 2005.

\bibitem{chen2020generative}
Mark Chen, Alec Radford, Rewon Child, Jeffrey Wu, Heewoo Jun, David Luan, and
  Ilya Sutskever.
\newblock Generative pretraining from pixels.
\newblock In {\em ICML}, 2020.

\bibitem{chen2021empirical}
Xinlei Chen, Saining Xie, and Kaiming He.
\newblock An empirical study of training self-supervised vision transformers.
\newblock In {\em ICCV}, 2021.

\bibitem{chen2021visformer}
Zhengsu Chen, Lingxi Xie, Jianwei Niu, Xuefeng Liu, Longhui Wei, and Qi~Tian.
\newblock Visformer: The vision-friendly transformer.
\newblock In {\em ICCV}, 2021.

\bibitem{Cordonnier2020relationship}
Jean-Baptiste Cordonnier, Andreas Loukas, and Martin Jaggi.
\newblock On the relationship between self-attention and convolutional layers.
\newblock {\em ICLR}, 2020.

\bibitem{Cubuk2018}
Ekin~D Cubuk, Barret Zoph, Dandelion Mane, Vijay Vasudevan, and Quoc~V Le.
\newblock {AutoAugment}: Learning augmentation policies from data.
\newblock In {\em CVPR}, 2019.

\bibitem{dai2020fbnetv3}
Xiaoliang Dai, Alvin Wan, Peizhao Zhang, Bichen Wu, Zijian He, Zhen Wei, Kan
  Chen, Yuandong Tian, Matthew Yu, Peter Vajda, et~al.
\newblock {FBNetV3}: Joint architecture-recipe search using neural acquisition
  function.
\newblock {\em arXiv:2006.02049}, 2020.

\bibitem{dascoli2021convit}
St{\'e}phane d'Ascoli, Hugo Touvron, Matthew Leavitt, Ari Morcos, Giulio
  Biroli, and Levent Sagun.
\newblock {ConViT}: Improving vision transformers with soft convolutional
  inductive biases.
\newblock In {\em ICML}, 2021.

\bibitem{Deng2009}
Jia Deng, Wei Dong, Richard Socher, Li-Jia Li, Kai Li, and Li~Fei-Fei.
\newblock {ImageNet}: A large-scale hierarchical image database.
\newblock In {\em CVPR}, 2009.

\bibitem{devlin2019bert}
Jacob Devlin, Ming-Wei Chang, Kenton Lee, and Kristina Toutanova.
\newblock {BERT}: Pre-training of deep bidirectional transformers for language
  understanding.
\newblock In {\em {NACCL}}, 2019.

\bibitem{dollar2021fast}
Piotr Doll{\'a}r, Mannat Singh, and Ross Girshick.
\newblock Fast and accurate model scaling.
\newblock In {\em CVPR}, 2021.

\bibitem{Dosovitskiy2020image}
Alexey Dosovitskiy, Lucas Beyer, Alexander Kolesnikov, Dirk Weissenborn,
  Xiaohua Zhai, Thomas Unterthiner, Mostafa Dehghani, Matthias Minderer, Georg
  Heigold, Sylvain Gelly, et~al.
\newblock An image is worth 16x16 words: Transformers for image recognition at
  scale.
\newblock In {\em ICLR}, 2021.

\bibitem{fan2021multiscale}
Haoqi Fan, Bo~Xiong, Karttikeya Mangalam, Yanghao Li, Zhicheng Yan, Jitendra
  Malik, and Christoph Feichtenhofer.
\newblock Multiscale vision transformers.
\newblock In {\em ICCV}, 2021.

\bibitem{Fukushima1980neocognitron}
Kunihiko Fukushima.
\newblock Neocognitron: A self-organizing neural network model for a mechanism
  of pattern recognition unaffected by shift in position.
\newblock {\em Biological cybernetics}, 36(4):193--202, 1980.

\bibitem{Goyal2017}
Priya Goyal, Piotr Doll{\'a}r, Ross Girshick, Pieter Noordhuis, Lukasz
  Wesolowski, Aapo Kyrola, Andrew Tulloch, Yangqing Jia, and Kaiming He.
\newblock Accurate, large minibatch {SGD}: Training {ImageNet} in 1 hour.
\newblock {\em arXiv:1706.02677}, 2017.

\bibitem{graham2021levit}
Ben Graham, Alaaeldin El-Nouby, Hugo Touvron, Pierre Stock, Armand Joulin,
  Herv{\'e} J{\'e}gou, and Matthijs Douze.
\newblock {LeViT}: a vision transformer in {ConvNet}'s clothing for faster
  inference.
\newblock In {\em ICCV}, 2021.

\bibitem{He2017}
Kaiming He, Georgia Gkioxari, Piotr Doll{\'a}r, and Ross Girshick.
\newblock {Mask R-CNN}.
\newblock In {\em ICCV}, 2017.

\bibitem{He2016}
Kaiming He, Xiangyu Zhang, Shaoqing Ren, and Jian Sun.
\newblock Deep residual learning for image recognition.
\newblock In {\em CVPR}, 2016.

\bibitem{hoffer2019augment}
Elad Hoffer, Tal Ben-Nun, Itay Hubara, Niv Giladi, Torsten Hoefler, and Daniel
  Soudry.
\newblock Augment your batch: better training with larger batches.
\newblock {\em arXiv:1901.09335}, 2019.

\bibitem{Ioffe2015}
Sergey Ioffe and Christian Szegedy.
\newblock Batch normalization: Accelerating deep network training by reducing
  internal covariate shift.
\newblock In {\em ICML}, 2015.

\bibitem{kolesnikov2019big}
Alexander Kolesnikov, Lucas Beyer, Xiaohua Zhai, Joan Puigcerver, Jessica Yung,
  Sylvain Gelly, and Neil Houlsby.
\newblock Big {Transfer} ({BiT}): General visual representation learning.
\newblock In {\em ECCV}, 2020.

\bibitem{Krizhevsky2012}
Alex Krizhevsky, Ilya Sutskever, and Geoffrey~E Hinton.
\newblock {ImageNet} classification with deep convolutional neural networks.
\newblock In {\em NeurIPS}, 2012.

\bibitem{LeCun1989}
Yann LeCun, Bernhard Boser, John~S Denker, Donnie Henderson, Richard~E Howard,
  Wayne Hubbard, and Lawrence~D Jackel.
\newblock Backpropagation applied to handwritten zip code recognition.
\newblock {\em Neural computation}, 1989.

\bibitem{liu2021swin}
Ze~Liu, Yutong Lin, Yue Cao, Han Hu, Yixuan Wei, Zheng Zhang, Stephen Lin, and
  Baining Guo.
\newblock Swin transformer: Hierarchical vision transformer using shifted
  windows.
\newblock In {\em ICCV}, 2021.

\bibitem{Long2015fully}
Jonathan Long, Evan Shelhamer, and Trevor Darrell.
\newblock Fully convolutional networks for semantic segmentation.
\newblock In {\em CVPR}, 2015.

\bibitem{Loshchilov2017decoupled}
Ilya Loshchilov and Frank Hutter.
\newblock Decoupled weight decay regularization.
\newblock In {\em ICLR}, 2019.

\bibitem{miller1995wordnet}
George~A Miller.
\newblock Wordnet: a lexical database for english.
\newblock {\em Communications of the ACM}, 1995.

\bibitem{Nair2010}
Vinod Nair and Geoffrey~E Hinton.
\newblock Rectified linear units improve restricted boltzmann machines.
\newblock In {\em ICML}, 2010.

\bibitem{Radosavovic2019}
Ilija Radosavovic, Justin Johnson, Saining Xie, Wan-Yen Lo, and Piotr
  Doll{\'a}r.
\newblock On network design spaces for visual recognition.
\newblock In {\em ICCV}, 2019.

\bibitem{Radosavovic2020}
Ilija Radosavovic, Raj~Prateek Kosaraju, Ross Girshick, Kaiming He, and Piotr
  Doll{\'a}r.
\newblock Designing network design spaces.
\newblock In {\em CVPR}, 2020.

\bibitem{Ramachandran2019stand}
Prajit Ramachandran, Niki Parmar, Ashish Vaswani, Irwan Bello, Anselm Levskaya,
  and Jonathon Shlens.
\newblock Stand-alone self-attention in vision models.
\newblock {\em NeurIPS}, 2019.

\bibitem{recht2019imagenet}
Benjamin Recht, Rebecca Roelofs, Ludwig Schmidt, and Vaishaal Shankar.
\newblock Do imagenet classifiers generalize to imagenet?
\newblock In {\em ICML}, 2019.

\bibitem{Ren2015}
Shaoqing Ren, Kaiming He, Ross Girshick, and Jian Sun.
\newblock Faster {R-CNN}: Towards real-time object detection with region
  proposal networks.
\newblock In {\em NeurIPS}, 2015.

\bibitem{ridnik2021imagenet}
Tal Ridnik, Emanuel Ben-Baruch, Asaf Noy, and Lihi Zelnik-Manor.
\newblock Imagenet-21k pretraining for the masses.
\newblock In {\em NeurIPS}, 2021.

\bibitem{Simonyan2015}
Karen Simonyan and Andrew Zisserman.
\newblock Very deep convolutional networks for large-scale image recognition.
\newblock In {\em ICLR}, 2015.

\bibitem{Szegedy2015}
Christian Szegedy, Wei Liu, Yangqing Jia, Pierre Sermanet, Scott Reed, Dragomir
  Anguelov, Dumitru Erhan, Vincent Vanhoucke, and Andrew Rabinovich.
\newblock Going deeper with convolutions.
\newblock In {\em CVPR}, 2015.

\bibitem{Szegedy2016a}
Christian Szegedy, Vincent Vanhoucke, Sergey Ioffe, Jonathon Shlens, and
  Zbigniew Wojna.
\newblock Rethinking the inception architecture for computer vision.
\newblock In {\em CVPR}, 2016.

\bibitem{Tan2019}
Mingxing Tan and Quoc~V Le.
\newblock {EfficientNet}: Rethinking model scaling for convolutional neural
  networks.
\newblock {\em ICML}, 2019.

\bibitem{tan2021efficientnetv2}
Mingxing Tan and Quoc~V Le.
\newblock Efficientnetv2: Smaller models and faster training.
\newblock In {\em ICML}, 2021.

\bibitem{Touvron2020training}
Hugo Touvron, Matthieu Cord, Matthijs Douze, Francisco Massa, Alexandre
  Sablayrolles, and Herv{\'e} J{\'e}gou.
\newblock Training data-efficient image transformers \& distillation through
  attention.
\newblock In {\em ICML}, 2021.

\bibitem{touvron2021going}
Hugo Touvron, Matthieu Cord, Alexandre Sablayrolles, Gabriel Synnaeve, and
  Herv{\'e} J{\'e}gou.
\newblock Going deeper with image transformers.
\newblock {\em arXiv:2103.17239}, 2021.

\bibitem{Vaswani2017attention}
Ashish Vaswani, Noam Shazeer, Niki Parmar, Jakob Uszkoreit, Llion Jones,
  Aidan~N Gomez, Lukasz Kaiser, and Illia Polosukhin.
\newblock Attention is all you need.
\newblock In {\em NeurIPS}, 2017.

\bibitem{wang2019learning}
Qiang Wang, Bei Li, Tong Xiao, Jingbo Zhu, Changliang Li, Derek~F Wong, and
  Lidia~S Chao.
\newblock Learning deep transformer models for machine translation.
\newblock In {\em {ACL}}, 2019.

\bibitem{wang2021pyramid}
Wenhai Wang, Enze Xie, Xiang Li, Deng-Ping Fan, Kaitao Song, Ding Liang, Tong
  Lu, Ping Luo, and Ling Shao.
\newblock Pyramid vision transformer: A versatile backbone for dense prediction
  without convolutions.
\newblock In {\em ICCV}, 2021.

\bibitem{Wang2018non}
Xiaolong Wang, Ross Girshick, Abhinav Gupta, and Kaiming He.
\newblock Non-local neural networks.
\newblock In {\em CVPR}, 2018.

\bibitem{wu2021cvt}
Haiping Wu, Bin Xiao, Noel Codella, Mengchen Liu, Xiyang Dai, Lu~Yuan, and Lei
  Zhang.
\newblock {CvT}: Introducing convolutions to vision transformers.
\newblock In {\em ICCV}, 2021.

\bibitem{Xie2017}
Saining Xie, Ross Girshick, Piotr Doll{\'a}r, Zhuowen Tu, and Kaiming He.
\newblock Aggregated residual transformations for deep neural networks.
\newblock In {\em CVPR}, 2017.

\bibitem{yuan2021incorporating}
Kun Yuan, Shaopeng Guo, Ziwei Liu, Aojun Zhou, Fengwei Yu, and Wei Wu.
\newblock Incorporating convolution designs into visual transformers.
\newblock In {\em ICCV}, 2021.

\bibitem{yuan2021tokens}
Li~Yuan, Yunpeng Chen, Tao Wang, Weihao Yu, Yujun Shi, Zihang Jiang, Francis~EH
  Tay, Jiashi Feng, and Shuicheng Yan.
\newblock Tokens-to-token {ViT}: Training vision transformers from scratch on
  {ImageNet}.
\newblock In {\em ICCV}, 2021.

\bibitem{yun2019cutmix}
Sangdoo Yun, Dongyoon Han, Seong~Joon Oh, Sanghyuk Chun, Junsuk Choe, and
  Youngjoon Yoo.
\newblock {CutMix}: Regularization strategy to train strong classifiers with
  localizable features.
\newblock In {\em CVPR}, 2019.

\bibitem{Zhang2018mixup}
Hongyi Zhang, Moustapha Ciss{\'{e}}, Yann~N. Dauphin, and David Lopez{-}Paz.
\newblock Mixup: Beyond empirical risk minimization.
\newblock In {\em ICLR}, 2018.

\bibitem{Zhao2020exploring}
Hengshuang Zhao, Jiaya Jia, and Vladlen Koltun.
\newblock Exploring self-attention for image recognition.
\newblock In {\em CVPR}, 2020.

\bibitem{zhong2020random}
Zhun Zhong, Liang Zheng, Guoliang Kang, Shaozi Li, and Yi~Yang.
\newblock Random erasing data augmentation.
\newblock In {\em AAAI}, 2020.

\end{thebibliography}


\end{document}